\documentclass{article}

    \usepackage[preprint]{neurips_2025}

\usepackage[utf8]{inputenc} 
\usepackage[T1]{fontenc}    
\usepackage{hyperref}       
\usepackage{url}            
\usepackage{booktabs}       
\usepackage{amsfonts}       
\usepackage{nicefrac}       
\usepackage{microtype}      
\usepackage{xcolor}         
\usepackage{graphicx}
\usepackage{amsmath}
\usepackage{multirow,booktabs,hhline}

\usepackage{times}
\usepackage{latexsym}

\usepackage[utf8]{inputenc} 
\usepackage[T1]{fontenc}    
\usepackage{hyperref}       
\usepackage{url}            
\usepackage{booktabs}       
\usepackage{amsfonts}       
\usepackage{nicefrac}       
\usepackage{epsfig,graphicx,subfigure}
\usepackage[algo2e]{algorithm2e}
\usepackage{bm}
\usepackage{wrapfig}
\usepackage{lipsum}
\usepackage{footnote}
\usepackage{bbding}
\usepackage[bottom]{footmisc}
\usepackage[toc,page]{appendix}
\usepackage{color}         
\usepackage{subfigure}

\usepackage[T1]{fontenc}
\usepackage[utf8]{inputenc}
\usepackage{authblk}
\usepackage{setspace}

\title{Low-hallucination Synthetic Captions for Large-Scale Vision-Language Model Pre-training}

\author[]{Xinsong Zhang\thanks{equal contribution}~~\thanks{Corresponding to: Xinsong Zhang <bertsyzhang@tencent.com>}~~}
\author[]{Yarong Zeng$^{\ast}$}
\author[]{Xinting Huang}
\author[]{\\ Hu Hu}
\author[]{Runquan Xie}
\author[]{Han Hu}
\author[]{Zhanhui Kang}
\affil[]{Hunyuan Team, Tencent}

\newcommand{\ModelName}{Hunyuan-Recap100M}

\begin{document}

\maketitle

\begin{abstract}
In recent years, the field of vision-language model pre-training has experienced rapid advancements, driven primarily by the continuous enhancement of textual capabilities in large language models. However, existing training paradigms for multimodal large language models heavily rely on high-quality image-text pairs. As models and data scales grow exponentially, the availability of such meticulously curated data has become increasingly scarce and saturated, thereby severely limiting further advancements in this domain. This study investigates scalable caption generation techniques for vision-language model pre-training and demonstrates that large-scale low-hallucination synthetic captions can serve dual purposes: 1) acting as a viable alternative to real-world data for pre-training paradigms and 2) achieving superior performance enhancement when integrated into vision-language models through empirical validation. This paper presents three key contributions: 1) a novel pipeline for generating high-quality, low-hallucination, and knowledge-rich synthetic captions. Our continuous DPO methodology yields remarkable results in reducing hallucinations. Specifically, the non-hallucination caption rate on a held-out test set increases from $48.3\%$ to $77.9\%$ for a 7B-size model. 2) Comprehensive empirical validation reveals that our synthetic captions confer superior pre-training advantages over their counterparts. Across 15 vision language tasks, the model trained with our data achieves a significant performance gain of at least $6.2\%$ compared to identical images with alt-text. In 20 common cognitive domains, the model trained with our data outperforms the alt-text data by at least $7.5\%$. 
Meanwhile, it also offers considerable support in the text-to-image domain. With the support of our dataset, the FID score is reduced by $17.1$ on a real-world validation benchmark and $13.3$ on the MSCOCO validation benchmark.
3) We will release {\bf \ModelName}, a low-hallucination and knowledge-intensive synthetic caption dataset, which will enable breakthroughs in vision language pre-training, cross-modal generation, and related applications.

\end{abstract}
\section{Introduction}
\label{sec:intro}

With the enormous power of large language models, remarkable performance gains have recently been achieved in a variety of tasks in natural language processing (NLP), computer vision (CV), and also in cross-modal fields~\citep{brown2020language,chung2022scaling,chowdhery2022palm,touvron2023llama,dosovitskiy2020image,alayrac2022flamingo,liu2023visual,achiam2023gpt,wang2024qwen2,hurst2024gpt}. The Large Language Model (LLM) and the Vision Language Model (VLM) are usually equipped with Transformer~\citep{vaswani2017attention} as the backbone and then pre-trained with a tremendous amount of unlabeled data. The strong representation ability of the model, the massive amount of data, and the effective means of training make the foundation models powerful for successfully solving the tasks of vision and language. 

While text data has experienced exponential growth and fuels large language models, vision language model pre-training confronts a unique data bottleneck: high-quality text-image alignments remain scarce despite their critical role in training vision language models. Previous approaches to building large-scale multimodal datasets rely on Web-based alt-text extraction~\citep{sharma2018conceptual,changpinyo2021conceptual,DBLP:conf/nips/OrdonezKB11,schuhmann2021laion,schuhmann2022laion,kakaobrain2022coyo-700m,gadre2023datacomp,liu2022taisu,gu2022wukong}. The alt-text data undergo a lengthy processing pipeline, involving stages such as text cleaning, image filtering, and cross-modal relevance validation, ultimately producing image-text pairs. Although significant data loss occurs at each processing stage and even with these efforts, the final output still suffers from inherent limitations such as low informativeness and partial misalignment between visual content and the corresponding descriptions. As naturally occurring high-quality image-text pairs are increasingly exhausted, traditional extraction methods experience considerable declines in output quality when scaled up, leading to significantly diminishing marginal returns in the training of VLMs. Therefore, novel techniques are urgently needed to produce large-scale, high-quality multimodal datasets. 

\begin{figure*}[ht]
\begin{center}
\centerline{\includegraphics[width=0.95\textwidth]{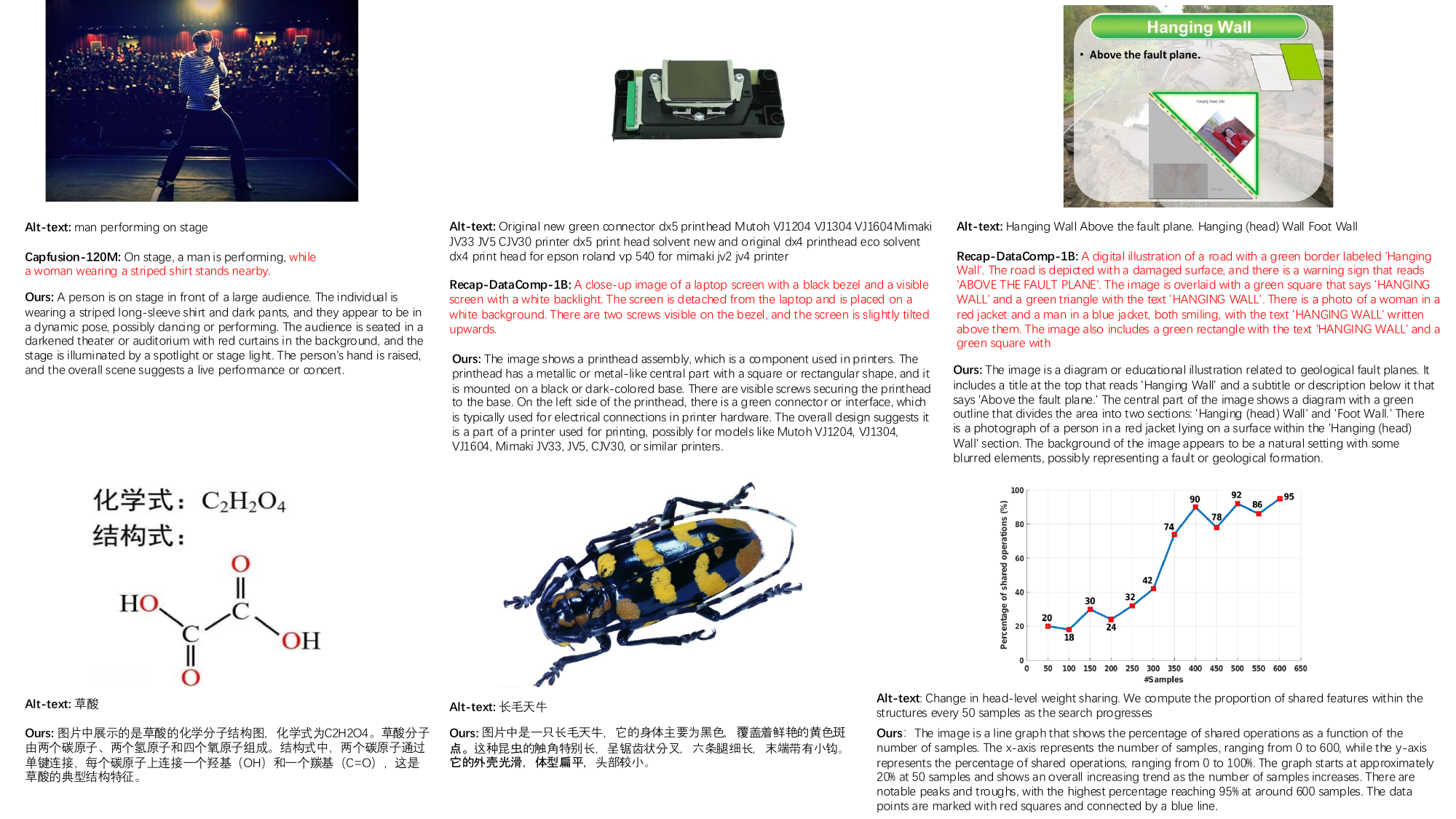}}
\caption{{\bf Examples of the original caption, previous synthetic caption and our synthetic caption.}}
\label{Fig:case}
\end{center}
\end{figure*}

Recently, several exploratory investigations have been carried out in multi-modal synthetic data, which have, to some extent, demonstrated their positive impact on various cross-modal tasks~\citep{li2023blip,betker2023improving,yu2024capsfusion,li2024if}. However, there are significant defects in all of them, mainly reflected in the presence of hallucinations and low information density. As shown in Figure~\ref{Fig:case}, there are a large number of hallucinations (labeled in red) in the previous synthetic captions, and the captions are rather monotonous, lacking informative expressions. Therefore, in this paper, we focus on addressing two issues: 1. How to establish an effective pipeline for generating low-hallucination synthetic captions? 2. How can more valuable information be injected into synthetic captions to increase the information density of synthetic data?

{\bf How to establish an effective pipeline for generating low-hallucination synthetic captions?} VLMs enable efficient large-scale caption synthesis, but persistent hallucination effects constrain their reliability~\citep{bai2024hallucination}. As shown in Figure~\ref{Fig:case}, Recap-DataComp-1B~\citep{li2024if} contains a significant amount of hallucinatory information.
Excess hallucinatory information in the training data is ultimately encoded in the final model, hindering the learning of correct patterns.
The severity of this problem is quantitatively demonstrated in Section~\ref{sec:experiments}.
Given the propensity for models trained on imperfect data to produce unfounded descriptions, effectively combating the generation of hallucinatory captions remains a paramount and often unavoidable challenge in caption synthesis.
To address the crucial issue of hallucination in synthesized captions, we propose employing the Direct Preference Optimization (DPO) method. This approach aims to mitigate hallucinations by increasing the model's confidence in generating descriptions of real visual content while simultaneously suppressing the confidence associated with hallucinatory visual elements. 
Furthermore, within the context of the synthetic caption task, we observe that DPO adheres to a significant scaling law within a specific data-size range. Notably, upon saturation of this scaling behavior, resampling of the data can further enhance the quality of the synthetic captions.
As demonstrated in Table~\ref{tab:data}, applying our proposed method (more detailed in Section~\ref{sec:model_description}) to the same dataset yields a significant improvement in the non-hallucination rate (evaluated using metrics described in Section~\ref{sec:eval_metric}). Specifically, our approach increases the non-hallucination rate from 29.7\% to 73.3\%, representing a substantial enhancement compared to the baseline approach used in Recap-DataComp-1B~\citep{li2024if}.

\begin{table}[ht]
\small
\centering	
\resizebox{0.90\columnwidth}{!}{
\begin{tabular}	{ c | c | c | c | c }
\toprule
Dataset & \# Length & \# Details $\uparrow$ & \# None Hallucinations $\uparrow$ & \# Hallucination Rate $\downarrow$ \\
\midrule
Capfusion-120M & 23.5 & 2.58 & 72.3\% & 14\% \\
Our Method & {\bf 80.4} & {\bf 6.9} & {\bf 73.9\%} & {\bf 5.4\%} \\
\midrule
Recap-DataComp-1B & 49.41 & {\bf 6.95} & 29.7\% & 24.9\% \\
Our Method & {\bf 72.60} & 6.64 & {\bf 73.3\%} & {\bf 5.64\%} \\
\midrule
{\ModelName} & {\bf 103.15} & {\bf 8.1} & {\bf 77.9\%} & {\bf 4.2\%} \\
 \bottomrule
\end{tabular}
}
\vspace{0.2cm}
\caption
{
{\bf A comparison between existing synthetic caption datasets and our proposed dataset in terms of hallucinations.} Here, "Length" represents the average length of the caption, "Details" represents the average number of visual details that appear in each caption, "None Hallucinations" indicates the proportion of non-hallucinatory data in the overall data, and "hallucination rate" represents the proportion of hallucinations among all details. A total of 1,000 data samples were taken for validation in each dataset.
}
\label{tab:data}
\end{table}

{\bf How can more valuable information be injected into synthetic captions to increase the information density of synthetic data?} 
VLM outputs often prioritize visual saliency, resulting in captions that are generally correct but informatively sparse, lacking deeper context and specific details derived from external or implicit knowledge. To overcome this limitation and produce synthetic captions with high information density, our approach involves training a VLM to effectively synthesize both external textual knowledge (e.g. alt-text, metadata) and its inherent world knowledge. Specifically, we used a powerful VLM, GPT-4o, guided by a meticulously designed prompt, to create the initial set of labeled data. A key aspect of this data generation process is the focused extraction and integration of vital information embedded in external text sources, which serves as a readily available source of descriptive and often factual information about images. For example, particular attention is paid to identifying and incorporating specific entity names or critical attributes within the alt-text, illustrated here by the case of oxalic acid in Figure~\ref{Fig:case}. Subsequently, the generated data undergoes a rigorous manual post-processing step to rectify obvious errors, confirm the inclusion of essential key information targeted during the generation phase, and effectively filter out noise or irrelevant details introduced during the initial knowledge injection. Finally, an open-source VLM, Qwen2-VL-7B, is fine-tuned on this carefully curated and information-rich dataset. 

Building upon the meticulously designed data generation and model training pipeline described previously, we developed and trained a specialized recaption model. 
We then proceeded to generate a large-scale synthetic dataset, {\ModelName}, comprising 100 million image-text pairs. 
Compared to models pre-trained using raw alt-text data or conventional synthetic captions, VLMs trained with the {\ModelName} dataset exhibit substantially improved overall performance. This significant performance gain, particularly pronounced in tasks requiring fine-grained visual perception, highlights the critical impact of training on synthetic data that not only offers high information density (as discussed previously), but also incorporates significantly reduced hallucinatory content, thereby enhancing the robustness and factual grounding of the downstream models. Empirical tests show models trained with our data gain at least 6.2\% on 15 VL tasks and 6.8\% on 20 cognitive domains compared to alternatives. This data also benefits text-to-image generation, reducing FID on validation benchmarks by up to 17.1.


Our contributions are summarized as follows:

1. {\bf High-Quality Synthetic Data Generation}: We develop a novel pipeline leveraging specialized methods to generate synthetic captions that are both significantly low in hallucination (improving the hallucination-free rate from $48.3\%$ to a notable $77.9\%$) and rich in knowledge.

2. {\bf Large-Scale Dataset and Demonstrated Impact}: We construct and will release {\ModelName}, a substantial 100 million image-text dataset generated by our pipeline. We demonstrate that pre-training vision-language models with {\ModelName} leads to substantially improved overall performance compared to existing synthetic captions, particularly enhancing perceptual capabilities, thus providing a valuable resource to the community.




\section{Related Work}
\label{sec:related}

{\bf Vision-Language Model.} 
Vision Language Models (VLMs) can be broadly classified into two categories. The first utilizes dual encoders, typically comprising a vision and a text encoder. These models are trained contrastively to learn aligned cross-modal representations and are highly effective in tasks such as image-text retrieval and zero-shot classification. Prominent examples include CLIP~\citep{radford2021learning}, FLIP~\citep{li2023scaling}, and SigLIP~\citep{zhai2023sigmoid}. The second category focuses on integrating visual information with a Large Language Model (LLM). Driven by the remarkable success of LLMs~\citep{brown2020language,openai2023gpt4,grattafiori2024llama,guo2025deepseek}, endowing these models with visual understanding has become a significant research direction. Flamingo~\citep{alayrac2022flamingo} pioneered large-scale VLM research by aligning a visual encoder with an LLM. LLaVA~\citep{liu2023visual} further established a methodology for training large multimodal models through instruction tuning. Subsequent efforts have continued to refine specific modules and optimize the overall performance of VLMs~\citep{wang2024qwen2,wu2024deepseek,mckinzie2024mm1,chen2024internvl,liu2024deepseek}.
Although current LLM-based vision language models demonstrate significantly greater capabilities compared to earlier multimodal models~\citep{zeng2021multi,singh2021flava,diao2022prefix,wang2022image,zhang2023toward}, they fundamentally share a core dependency with dual encoder VLMs (the first category discussed previously): the need for large-scale image-text pairs for training. This reliance presents a growing challenge, however, as the supply of high-quality image-text pair data is finite~\citep{villalobos2022will} and its expansion struggles to keep pace with the increasing scale of models. To overcome this data bottleneck, synthetic data offers a promising potential solution.

{\bf Synthetic data.} Building on its success in enhancing large language models~\citep{zhang2024automathtext,shao2024deepseekmath,zhu2024deepseekcoderv2,yang2024qwen2p5}, synthetic data generation is increasingly being explored within the multimodal domain. Some methods, such as BLIP-2~\citep{li2023blip}, utilize image captioning models to synthesize short captions as substitutes for alt-text, with CapFusion~\citep{yu2024capsfusion} leveraging LLMs for refinement. However, captions generated by these approaches often remain simplistic, providing minimal added information value beyond the original alt-text. Recap-DataComp-1B~\citep{li2024if} attempts to transform alt-text using existing vision language models, but faces the significant challenge of hallucination inherent in current VLMs. Other works focusing on dense captioning, including Allava~\citep{chen2024allava} and Dense-Fusion-1M~\citep{li2024densefusion}, generate localized descriptions, but the resulting datasets can also suffer from notable hallucinations. These limitations in existing synthetic VLM data highlight the need for more sophisticated generation techniques.
In summary, generating high-quality multimodal synthetic data faces challenges related to knowledge enrichment and hallucination control. Although VLM hallucination is a studied topic~\citep{bai2024hallucination}, a comprehensive synthetic visual captioning method capable of simultaneously producing diverse and rich content with minimal hallucination is still needed.

Our approach makes substantial progress in the generation of synthetic visual captions with improved knowledge injection and significantly reduced hallucination. This yields a caption dataset whose efficacy is demonstrated for large-scale VLM training, offering a viable alternative that overcomes the constraints of limited real data and the quality challenges of scaling traditional datasets.
\section{Methodology}
\label{sec:model_description}

In this section, we present our synthetic data pipeline, including continuous DPO aimed at reducing hallucinations, knowledge-enriching SFT to enhance knowledge content, and details of the model training pipeline and hyper-parameters.

\subsection{Continuous DPO}
\label{sec:method_cdpo}

Reinforcement learning has been reported to be an effective method of mitigating hallucinations in large language models~\citep{zhang2023hallucination,huang2025survey}. Direct Preference Optimization (DPO)~\citep{rafailov2023direct} emerges as a reinforcement learning based fine-tuning paradigm for LLM that directly optimizes preference data, effectively mitigating harmful content generation, such as hallucinations. Consequently, the application of DPO to mitigate hallucinations in image captions appears to be a natural and logical approach. The optimization objectives of the DPO algorithm are shown in Equation~\ref{equ:dpo1} and Equation~\ref{equ:dpo2}. As noted in~\cite{rafailov2023direct}, the loss function gradient $\mathcal{L}_{\text{DPO}}$ increases the probability of preferred completions $y_w$ while simultaneously reducing the probability of dispreferred completions $y_l$. Intuitively, by maximizing the probability of samples free of hallucinations and minimizing the probability of samples containing hallucinations, our recaption model can generate captions that are as free from uncertainty and potential hallucinations as possible.

\begin{equation}
\label{equ:dpo1}
\mathcal{L}_{\text{DPO}} = -\mathbb{E}_{(x,y_w,y_l)\sim \mathcal{D}} \left[ \log\sigma\left(\beta\log\frac{\pi_{\theta}(y_w|x)}{\pi_{ref}(y_w|x)} - \beta\log\frac{\pi_{\theta}(y_l|x)}{\pi_{\text{ref}}(y_l|x)} \right) \right]
\end{equation}

\begin{equation}
\label{equ:dpo2}
\begin{aligned}
&\nabla_{\theta} \mathcal{L}_{\text{DPO}}(\pi_{\theta}; \pi_{\text{ref}}) = 
\\
& - \beta \mathbb{E}_{(x, y_w, y_l) \sim \mathcal{D}} \left[ \underbrace{\sigma(\hat{r}_{\theta}(x, y_l) - \hat{r}_{\theta}(x, y_w))}_{\text{higher weight when reward estimate is wrong}} \left[ \underbrace{
\nabla_{\theta} \log \pi(y_w \mid x)}_{\text{increase likelihood of } y_w} - \underbrace{
\nabla_{\theta} \log \pi(y_l \mid x)}_{\text{decrease likelihood of } y_l} \right] \right],
\end{aligned}
\end{equation}

where $\hat{r}_{\theta}(x, y) = \beta \log \frac{\pi_{\theta}(y \mid x)}{\pi_{\text{ref}}(y \mid x)}$ is the reward implicitly defined by the language model $\pi_{\theta}$ and the reference model $\pi_{\text{ref}}$.

However, in our task, standard DPO does not exhibit continuous performance improvement with respect to data scale. Specifically, after initial rapid performance gains, its performance plateaus, as illustrated in Figure~\ref{exp:dpo_scale}. Examining the DPO objective (Equation~\ref{equ:dpo2}), we see that the training examples are weighted based on the implicit reward derived from the reference model, which reflects the preference strength between chosen and rejected completions. As the scale of the DPO dataset increases, the implicit reward signal, derived from the fixed reference model, does not change significantly. Consequently, the optimization plateaus as the model converges to a local optimum defined by this relatively static signal. To address this limitation, we propose Continuous DPO (CDPO), a modification of the standard DPO algorithm. In CDPO, when performance plateaus, we iteratively update the reference model (e.g., using the current policy), resample preference data using the updated policy, which in turn yields a new implicit reward signal, and resume optimization. Furthermore, during CDPO training, to prevent the sample length from acting as an unintended learning signal (e.g. if longer completions are more likely to contain hallucinations), we apply a sequence-length balancing operation to the preference pairs. This ensures that the average lengths of the preferred and dispreferred completions within the dataset are closely matched.

\subsection{Knowledge-enriching SFT}
\label{sec:method_ksft}

Previous recaption models (e.g. \citep{li2023blip,yu2024capsfusion}) often produce generalized descriptions lacking specificity, such as "a man performing on stage" in Figure~\ref{Fig:case}. This type of descriptive information has limited informational value, thus offering little assistance in training vision language models. Hence, our work significantly emphasizes the integration of valuable information. We define valuable information as comprising two main aspects: the richness of descriptive details and the knowledge content of the description. The latter involves the incorporation of external knowledge related to the image. For example, considering "oxalic acid" in Figure~\ref{Fig:case}, we aim for descriptions such as: \emph{The image shows the molecular chemical structure of oxalic acid, with the chemical formula $C_2H_2O_4$. The oxalic acid molecule consists of two carbon atoms, two hydrogen atoms, and four oxygen atoms. In the structural formula, the two carbon atoms are connected by single bonds, and each carbon atom is attached to a hydroxyl group ($OH$) and a carbonyl group ($C=O$), which are the typical structural features of oxalic acid.}

\begin{figure*}[htbp]
\begin{center}
\centerline{\includegraphics[width=0.9\textwidth]{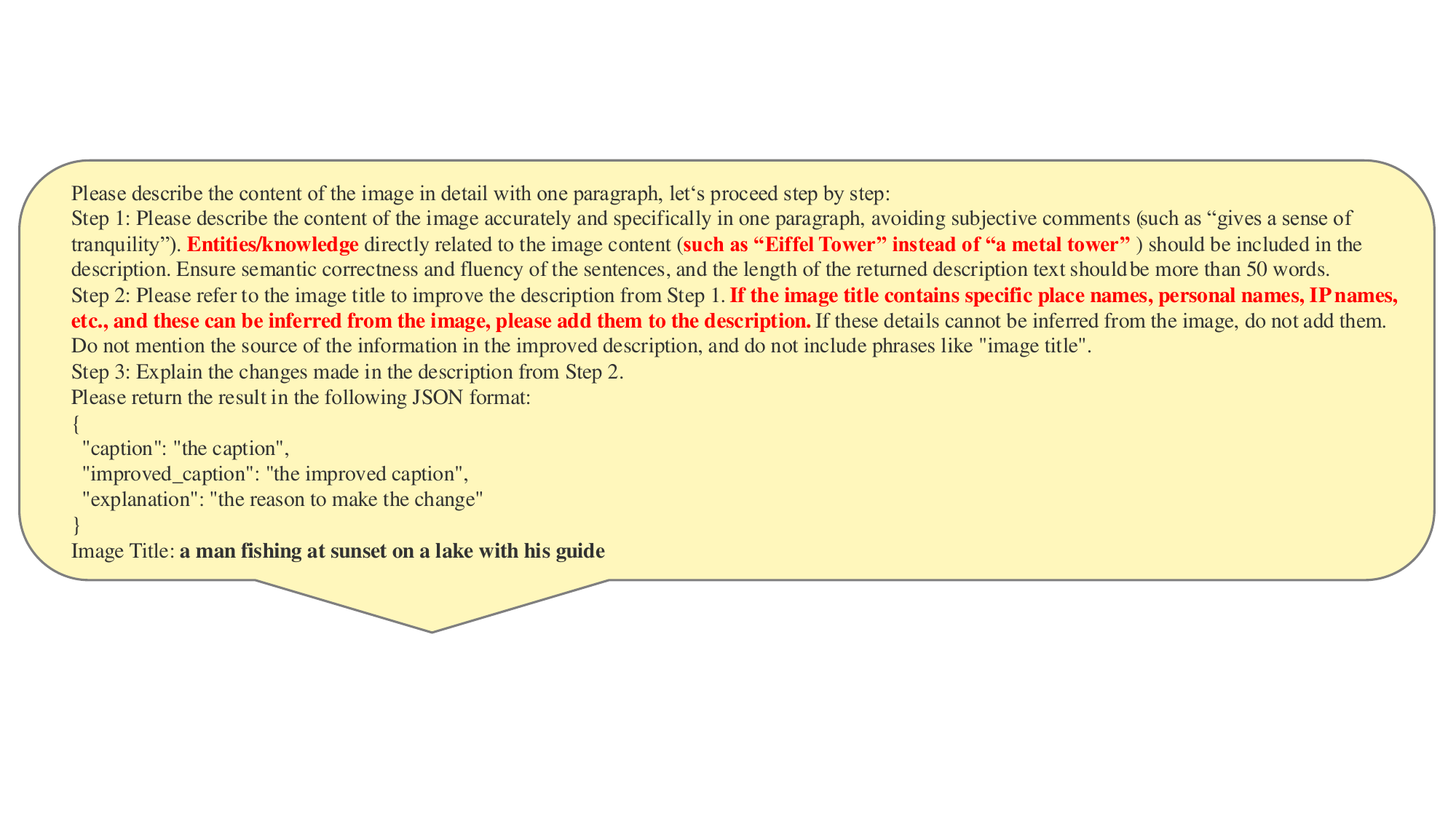}}
\vspace{0.2cm}
\caption{{\bf A GPT-4o Prompt for Knowledge-Rich Captions.} The portion highlighted in red within the prompt emphasizes the instructions for knowledge injection. This prompt utilizes a three-step procedure, which is intended to help ensure the inclusion of a rich body of world knowledge in the generated caption.}
\label{Fig:GPT-4o-prompt}
\end{center}
\end{figure*}

By constructing SFT data with captions rich in descriptive details, we enable the model, through fine-tuning, to learn to produce detailed and specific descriptions.
Therefore, we designed a detailed prompt and used GPT-4o to generate the initial training data. While GPT-4o understands the prompt requirements, the output data still contain a significant number of hallucinations. To address this, we conducted a manual review of the generated captions. By integrating these data processing steps, we constructed the knowledge-enhanced supervised fine-tuning data suitable for training models capable of generating rich image captions, such as the "arctolamia gestro" illustrated in Figure~\ref{Fig:case}.


\subsection{Training Pipeline}
\label{sec:method_train_pipeline}

\begin{figure*}[htbp]
\begin{center}
\centerline{\includegraphics[width=0.75\textwidth]{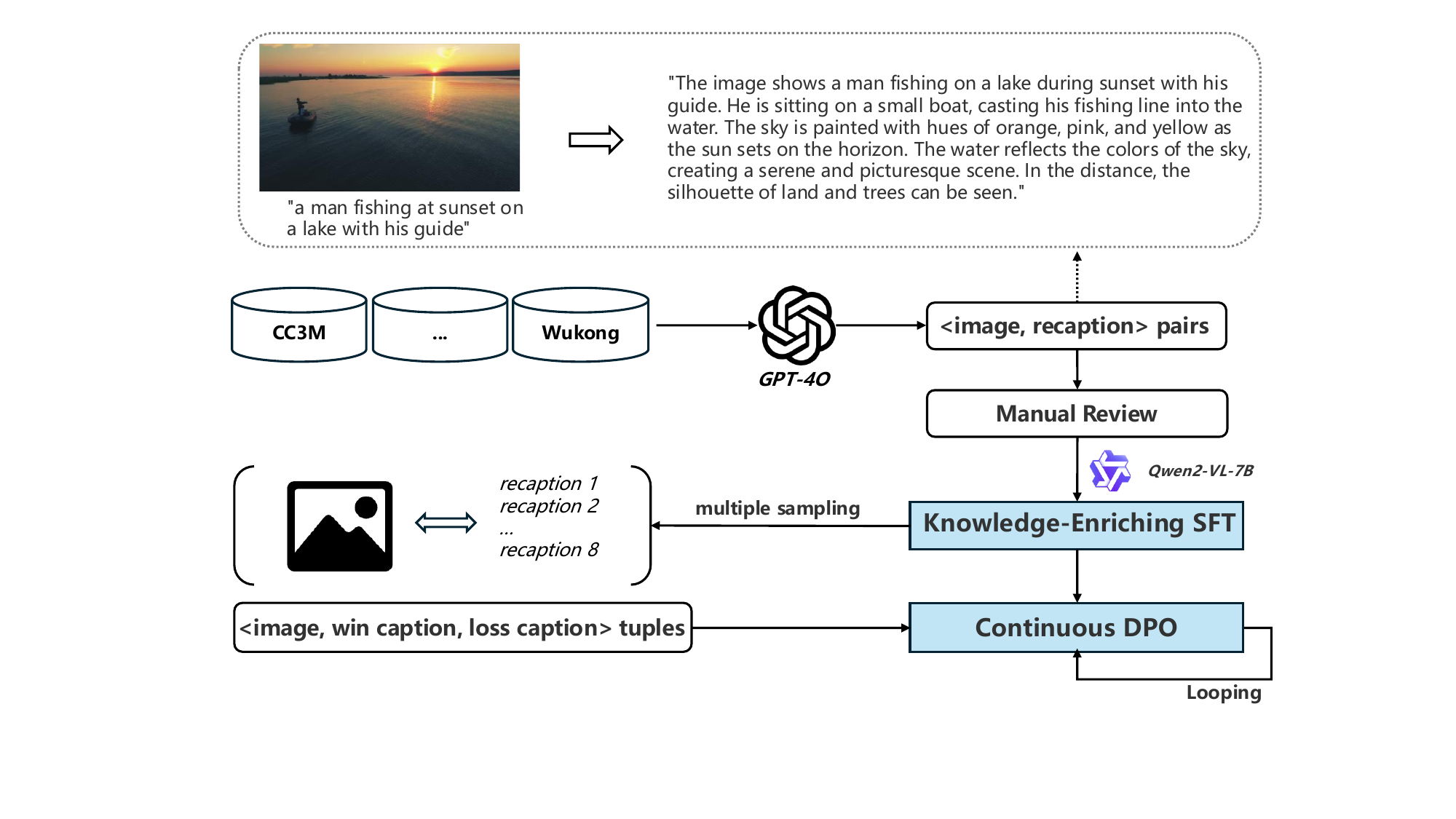}}
\vspace{0.2cm}
\caption{{\bf The illustration of our recaptioning pipeline.} The pipeline comprises the following key stages: Initial Data Generation with GPT-4o, Manual Review, Knowledge-Enriching SFT, and Continuous DPO.}
\label{Fig:train_pipeline}
\end{center}
\end{figure*}

As mentioned above, our comprehensive training pipeline consists of two sequential stages: knowledge-enriching SFT followed by continuous DPO. In this section, we introduce these stages specifically. The general pipeline of our work is shown in Figure~\ref{Fig:train_pipeline}. We adopt Qwen2-VL-7B~\citep{wang2024qwen2} as the base model and use the LLaMA Factory~\citep{zheng2024llamafactory} platform to carry out the overall training process.

{\bf Knowledge-enriching SFT.} We sampled raw image-text pairs from open-source multimodal datasets including CC3M~\citep{sharma2018conceptual}, CC12M~\citep{changpinyo2021conceptual}, DataComp~\citep{gadre2023datacomp}, Wukong~\citep{gu2022wukong}, and Wikipedia~\citep{srinivasan2021wit}. Using the prompt in Figure~\ref{Fig:GPT-4o-prompt}, we employed GPT-4o to generate the initial recaption data. Following a meticulous manual review process, $43,408$ data entries were retained. Subsequently, we performed LoRA fine-tuning on the Qwen2-VL-7B model. Specifically, training was conducted with a learning rate of $1.0e-5$, a global batch size of
$128$, and for $10$ epochs. This process yielded our Recaption-SFT model.


{\bf Continuous DPO.} As outlined in Section~\ref{sec:method_cdpo}, our methodology begins with training the initial DPO model. We first randomly sample $300,000$ image-text pairs and then employ the SFT model for inference. To generate diverse candidates, we used the following sampling hyperparameters: TOPP=$1.0$, TOPK=$20$, and TEMPERATURE=$1.0$. Each image-text pair underwent eight parallel inference iterations, yielding eight divergent candidates. We then constructed preference pairs (chosen/rejected pairs) by selecting the best and worst outputs from each set using an internal critic model.
~\footnote{This expert system evaluates caption quality by dissecting vision details in captions with an fine-tuned LLM (72B-sized) and assessing hallucination levels for each vision detail with a fine-tuned VLM (72B-sized).}.
To eliminate length-induced training bias, we implement length-balanced sampling that retains $218k$ high-quality pairs. This process ensures alignment in the length distribution between preferred and dispreferred completions. Subsequently, this curated data set was used for DPO training on the SFT model, yielding the initial DPO model (used as a reference model).
Utilizing the trained initial DPO model, we sampled additional $200k$ image-text pairs from the data sources. These pairs were used to generate new preference data employing the model itself, the same sampling hyperparameters, and the critic model. Applying the length-balancing strategy to filter these generated pairs, we retained $139k$ preferred/dispreferred pairs for the subsequent CDPO training iteration. All hyperparameters at different stages are listed in Table~\ref{tab:hyperparameters}.

\begin{table}[ht]
\small
\centering	
\resizebox{0.6\columnwidth}{!}{
\begin{tabular}	{ c | c | c | c }
\toprule
 & Stage-1-SFT & Stage-2-DPO & Stage-2.5-CDPO \\
\midrule
Resolution \# pixels & [3136, 12845056] & [3136, 12845056] & [3136, 12845056] \\
\midrule
Dataset \# Samples & 43K & 218K & 139K \\
\midrule
Finetuning Type & LoRA & LoRA & LoRA \\
\midrule
Batch Size & 128 & 64 & 64 \\
Learning Rate & $1e-5$ & $5e-6$ & $5e-6$ \\
Learning Epoch & 10 & 1 & 1 \\
 \bottomrule
\end{tabular}
}
\vspace{0.2cm}
\caption
{
{\bf Detailed configuration for each training stage of our recaption model.} The table outlines the progression of vision parameters, dataset characteristics, model specifications and training hyperparameters.
}
\label{tab:hyperparameters}
\end{table}
\section{Experiments}
\label{sec:experiments}

This section covers the quality assessment of our recaption data. We first outline the criteria used for this assessment, and then evaluate the data across two dimensions: its intrinsic quality and its impact on large-scale vision language model pre-training.

\begin{figure}[htbp]
	\centering
	\subfigure[Non-hallucination Rate] {\includegraphics[width=.45\textwidth]{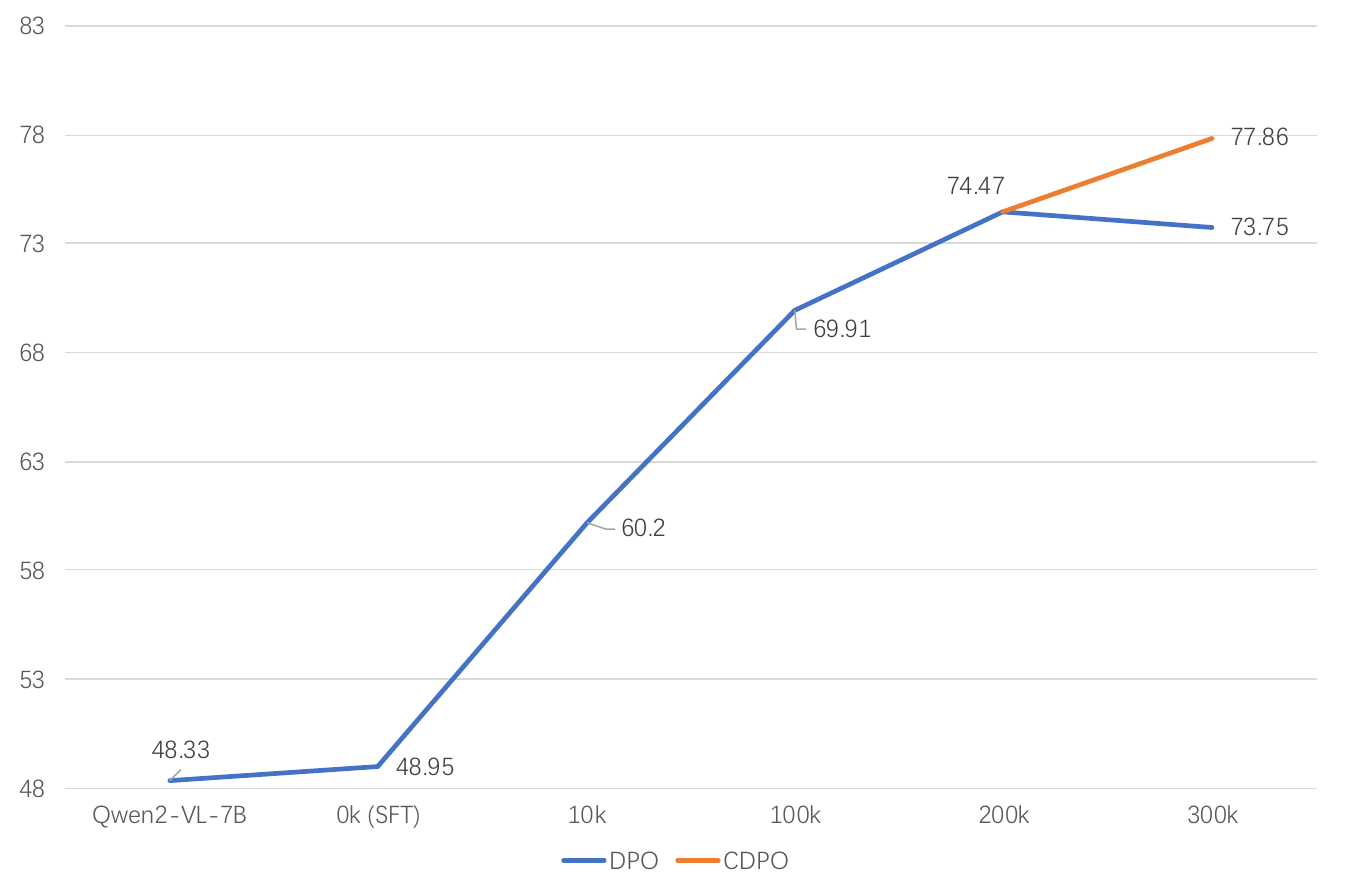}}
	\subfigure[Low-hallucination Rate] {\includegraphics[width=.45\textwidth]{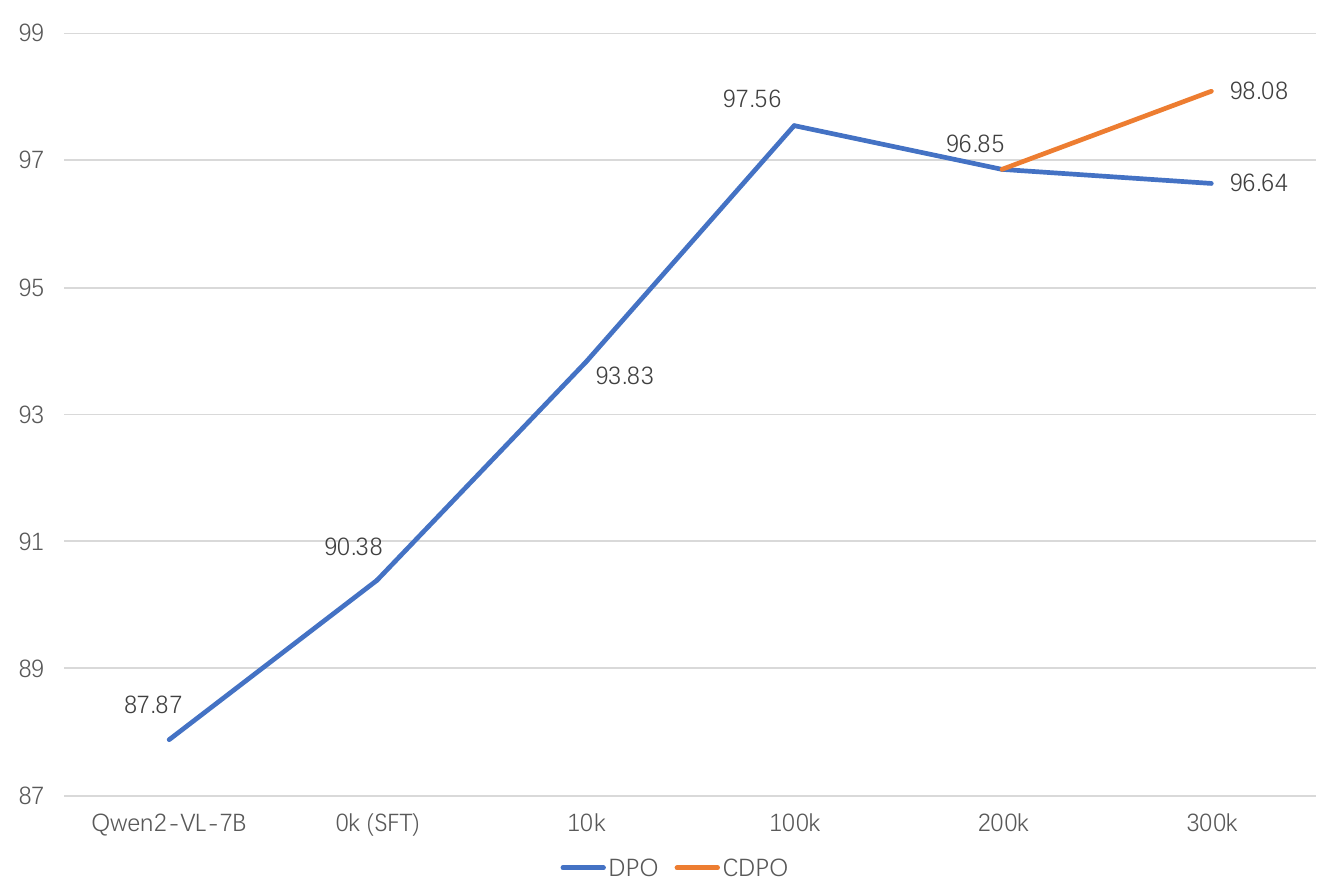}}
	\caption{{\bf Performance of different DPO strategies.} The horizontal axis represents the training data scale, while the vertical axis depicts the proportion of hallucination-free captions on the validation set (in percentage). Initially, the CDPO strategy utilizes the same training data as DPO. Upon reaching a training scale of $218k$ data points, it incorporates an additional $139k$ new data instances.}
	\label{exp:dpo_scale}
\end{figure}

\subsection{Evaluation Metrics}
\label{sec:eval_metric}

The quality of recaption data is assessed along two key dimensions. One dimension concerns the presence of harmful information, particularly hallucinations, which must be minimized. The other is content richness, whose value is validated by its effectiveness in model training. Consequently, our primary quality control efforts focus on mitigating hallucinations. The richness of the data, conversely, is evaluated through the training performance of large-scale vision language models, as detailed in Section~\ref{sec:perform_data}.
Adopting the hallucination assessment method from CIEM~\citep{hu2023ciem}, we utilize the large language model, GPT-4o, to parse each caption into its constituent visual details. GPT-4o then assesses each extracted detail for the presence of hallucinations. Following this detail-level judgment, we count the total number of hallucinatory details within a caption. Based on this count, we define two key metrics: the {\bf non-hallucination rate} as the proportion of captions containing no hallucinatory details and the {\bf low-hallucination rate} as the proportion of captions containing at most two hallucinatory details.

\subsection{Performance of Recaption Model}
\label{sec:perform_recap}

To evaluate our model, we apply it to the identical image-text pairs used by other recaption methods, such as Capfusion~\citep{yu2024capsfusion} and Recap-DataComp-1B~\citep{li2024if}. The evaluation results are presented in Table~\ref{tab:data}. When generating data for these same input pairs, our model produces synthetic captions with significantly more details than those generated by Capfusion and substantially fewer hallucinations than those from Recap-DataComp-1B.
Specifically, when evaluated on the same data, our model demonstrates substantial improvements over the baselines. Compared to Capfusion, our generated captions exhibit a $267\%$ increase in detail count, are $3.5$ times longer on average, and show a $8.6\%$ percentage point decrease in the rate of hallucinatory details. Relative to Recap-DataComp-1B, our model achieves a $43.6\%$ percentage point increase in the non-hallucination rate. It is important to note that while the brevity of Capfusion's captions leads to a relatively high non-hallucination rate, the limited detail content restricts their overall value. 

We then report the evaluation results of our recaption model on another open source dataset, Wukong~\citep{gu2022wukong}. As illustrated in Figure~\ref{exp:dpo_scale}, the base model (Qwen2-VL-7B) exhibits a significant hallucination problem in this dataset. Specifically, its non-hallucination rate is $48.33\%$ and its low-hallucination rate is $87.87\%$, figures that indicate a substantial level of hallucinatory content.
Following supervised fine-tuning, the low-hallucination rate is significantly increased. Subsequently, the application of DPO leads to substantial improvements in both the non-hallucination rate and the low-hallucination rate, increasing them $29.53\%$ and $10.21\%$, respectively. Furthermore, we observe that DPO exhibits a notable scaling law during the initial training stage for the recaption task; however, performance plateaus once the data size reaches a certain scale. Our CDPO algorithm provides further enhancement to both metrics, increasing the non-hallucination rate by $3.4\%$ and the low-hallucination rate by $0.5\%$.

\begin{table*}[ht]
\centering
\resizebox{0.98\textwidth}{!}{
\begin{tabular}{l|cccc|ccccc|cccc|cc|c}
\toprule
 & \multicolumn{4}{c}{Scientific QA} & \multicolumn{5}{c}{OCR QA} & \multicolumn{4}{c}{Benchmark QA}  & \multicolumn{2}{c}{Hallucination Tasks} & \\
 Dataset & MathVista & MMMU & AI2D & ScienceQA & OCRBench & DocVQA & TextVQA & InfographicsVQA & ChartQA & MMBench & MMStar & RealworldQA & MMVet & HallusionBench & None-Hallu. Rate & Avg. 
\\

 & testmini & val & test & test & test
 & val & val & val & test (aug./hum.) & en-dev & test & test & test & test & -- & 
\\

\midrule 
Alt-text(20M) & 42.20 & 42.24 & 67.06 & 81.75 & 49.0 & 65.01 & 55.97 & 31.17 & 46.92 & 67.96 & 44.27 & 57.12 & 43.81 & 34.29 & 68.38 & 53.12 \\
Recap-DataComp-1B(20M) & 51.40 & 43.33 & 68.60 & 86.27 & 55.8 & 46.90 & 64.75 & 33.52 & 53.16 & 66.58 & 48.46 & 58.95 & 39.66 & 29.84 & 33.79 & 52.07 \\
{Qwen2-VL-7B(20M)} & {\bf 54.40} & 42.33 & {\bf 73.77} & {\bf 87.31} & {\bf 61.3} & 62.88 & 71.9 & 40.92 & 56.68 & 69.24 & 48.86 & 59.74 & {\bf 48.20} & 31.84 & 65.13 & 58.31 \\
{\ModelName(20M)} & 52.50 & {\bf 44.27} & 73.59 & 85.82 & 57.3 & {\bf 66.27} & {\bf 72.59} & {\bf 42.47} & {\bf 57.00} & {\bf 72.77} & {\bf 50.13} & {\bf 60.26} & 46.02 & {\bf 35.45} & {\bf 73.87} & {\bf 59.35} \\
{\ModelName} & 45.10 & 43.56 & 73.19 & 85.92 & {\underline{65.80}} & {\underline{66.30}} & 72.42 & {\underline{44.66}} & 56.0 & {\underline{74.83}} & 49.13 & {\underline{64.44}} & {\underline{49.00}} & 34.60 & 72.61 & {\underline{59.83}} \\
\midrule 
Alt-text(20M)$\ast$ & 50.10 & 45.0 & 73.18 & 85.57 & 59.8 & 77.34 & 68.47 & 43.99 & 62.08 & 72.42 & 47.89 & 61.57 & 42.09 & 33.80 & 39.27 & 57.50 \\
Recap-DataComp-1B(20M)$\ast$ & 53.70 & 44.44 & 74.65 & 86.12 & 62.1 & 79.11 & 66.39 & 45.38 & 64.28 & 65.89 & 50.44 & 61.96 & 39.34 & 32.77 & 43.62 & 58.01 \\
{Qwen2-VL-7B(20M)}$\ast$ & 52 & {\bf 45.22} & 77.19 & 88.60 & {\bf 66.7} & {\bf 83.13} & 70.77 & {\bf 51.92} & 66.84 & {\bf 74.31} & 52.71 & {\bf 65.49} & {\bf 47.29} & 31.98 & 66.20 & 62.69 \\
{\ModelName(20M)}$\ast$ & {\bf 56.20} & 44.00 & {\bf 77.91} & {\bf 89.74} &  65.9 & 83.06 & {\bf 71.37} & 50.32 & {\bf 68.96} & 72.34 & {\bf 54.66} & {\bf 65.49} & 43.17 & {\bf 36.38} & {\bf 71.86} & {\bf 63.42} \\
{\ModelName}$\ast$ & 48.80 & {\underline{45.22}} & 77.45 & 87.90 & 65.7 & {\underline{83.99}} & {\underline{75.09}} & 51.83 & 65.76 & {\underline{74.31}} & 54.62 & {\underline{69.67}} & 45.17 & {\underline{36.45}} & 70.65 & {\underline{63.51}} \\
\bottomrule
\end{tabular}}
\vspace{0.2cm}
\caption{
\textbf{Experimental results on 15 vision-language tasks.} Avg. denotes the average score across all evaluated vision-language tasks.
The Non-Hallucination Rate is a hallucination assessment metric defined in Section~\ref{sec:eval_metric}. For evaluation, we randomly sampled 1000 instances from the DataCamp dataset to quantify the non-hallucination rate.
In the table, bold text indicates the optimal performance for a given metric under the $20M$ data setting, while underlined text signifies further enhancement achieved when utilizing the entire dataset.
The $\ast$ symbol indicates that a light, unified supervised fine-tuning (SFT)~\citep{wang2023cogvlm,li2024llava} was applied after pre-training.
It should be noted that our primary objective is to investigate the impact of different caption datasets on VLM model training. Consequently, these results represent comparative experiments conducted at a fixed data scale ($20M$). They were performed without employing advanced fine-tuning or reinforcement learning techniques often used to boost performance towards state-of-the-art, and thus are not directly comparable to SOTA models.}
\label{tab:main_result}
\end{table*}

\subsection{Performance of Recaption Data}
\label{sec:perform_data}

We validate the effectiveness of {\ModelName} in two scenarios, vision-language model pre-training and text-to-image generation. This validation is conducted primarily from the viewpoints of model training efficacy, hallucination levels, and knowledge coverage.

{\bf Vision-Language Model Pre-Training.} 
We first validate the effectiveness of our dataset for the pre-training of vision language models. This aims to demonstrate that, compared to existing alt-text and other recaption datasets, {\ModelName} can facilitate more effective training of vision language models, leading to superior performance on both general open-source multimodal tasks and our internal knowledge-related test sets. 
For the VLM framework, we adopt the LLaVA~\citep{liu2023visual} architecture, comprising SigLIP~\citep{DBLP:conf/iccv/ZhaiM0B23} as visual encoder and Hunyuan-7B~\citep{Hunyuan7B} as the base large language model. To ensure a fair comparison of the influence of caption datasets, we sample $20M$ data points from each of the alt-text and recaption datasets. Note that the experiments that use the alt-text data utilize the same image set as our {\ModelName} dataset. In contrast, the Recap-Datacamp-1B data uses a randomly sampled subset, implying a different image set distribution compared to the alt-text and {\ModelName} data.\footnote{{\ModelName} dataset is compiled from multiple open-source datasets, resulting in a distribution that is not exactly the same as that of DataCamp.} In Table~\ref{tab:main_result}, "Qwen2-VL-7B(20M)" refers to the model trained on the $20M$ recaption data generated by the unoptimized Qwen2-VL-7B model, which contains more hallucinations.
On 15 open-source multimodal tasks, the VLM model pre-trained with {\ModelName} data significantly outperforms models pre-trained with alt-text or other synthetic captions. This superiority is particularly evident in hallucination-related tasks, where training with {\ModelName} data achieves a substantial lead over training with other counterparts. Additionally, we monitored the loss function to evaluate the model convergence behavior in the appendix~\ref{appendix:loss}. When subjected to identical supervised fine-tuning datasets, models pre-trained with {\ModelName} demonstrated markedly improved convergence properties.

Furthermore, to quantify the enhancement in the visual perception capabilities of VLM attributable to {\ModelName}, we manually collected $1,197$ images with diverse visual characteristics across 20 domains, including animals, plants, clothing, and landmarks, among others. These images are accompanied by manually annotated labels for evaluation. We assessed the VLM's perceptive ability on this dataset, and the results are presented in Table~\ref{tab:object_result}.
When trained with a $20M$ subset of {\ModelName} data, the VLM model significantly outperforms other counterparts in 14 of the 20 categories, demonstrating an average improvement of at least $3.8\%$ compared to models trained on other $20M$ datasets. Upon utilizing the full {\ModelName} dataset for training, the model shows significant improvements in 18 of the 20 perception tasks, with an overall average improvement reaching $9\%$.

\begin{table*}[ht]
\centering
\resizebox{0.98\textwidth}{!}{
\begin{tabular}{l|cccccccccccccccccccc|c}
\toprule
 Dataset & Life & Landmark & Celebrity & Food & Cloth & Good & Game & Traffic & Art & Festival & Moive & Icon & Animal & Plant & Vegetables & Sport & Medical & Military & Tool & History & Avg. 
\\

\midrule 
Alt-text(20M) & {\bf 54.7} & 36.83 & 22.67 & 24.13 & 29.93 & 37.07 & 25.30 & 50.57 & 29.47 & 26.73 & 17.10 & 41.80 & 60.57 & 37.77 & 24.73 & 67.77 & {\bf 41.30} & {\bf 50.30} & 28.17 & 31.00 & 36.33 \\

Recap-DataComp-1B(20M) & 32.57 & 26.60 & 20.87 & 21.53 & 34.47 & 30.07 & 24.83 & 46.57 & 19.47 & 10.20 & 14.73 & 25.50 & 54.57 & 28.57 & 26.93 & 60.67 & 24.63 & 36.13 & 37.60 & 21.93 & 29.67 \\

{Qwen2-VL-7B(20M)} & 45.83 & 35.60 & {\bf 34.47} & 31.90 & {\bf 49.77} & 45.73 & {\bf 34.80} & 66.10 & 36.30 & 23.83 & 21.40 & 40.60 & 54.30 & 32.17 & 29.43 & 69.83 & 39.73 & 43.87 & 28.90 & {\bf 36.53} & 39.97 \\

{\ModelName(20M)} & 42.80 & {\bf 45.30} & 27.10 & {\bf 32.03} & 43.90 & {\bf 54.43} & {\bf 34.80} & {\bf 68.23} & {\bf 37.83} & {\bf 28.73} & {\bf 28.13} & {\bf 52.70} & {\bf 61.83} & {\bf 44.10} & {\bf 38.27} & {\bf 74.30} & 34.00 & 45.60 & {\bf 45.13} & 35.23 & {\bf 43.80}\\

{\ModelName} & 53.43 & {\underline{52.10}} & {\underline{39.03}} & {\underline{46.07}} & {\underline{58.70}} & {\underline{55.40}} & {\underline{52.53}} & {\underline{74.57}} & {\underline{44.57}} & {\underline{40.47}} & {\underline{33.13}} & {\underline{67.10}} & {\underline{75.27}} & {\underline{62.60}} & {\underline{54.07}} & {\underline{83.87}} & {\underline{41.50}} & {\underline{50.87}} & 41.63 & {\underline{43.53}} & {\underline{53.67}}\\

\bottomrule
\end{tabular}}
\vspace{0.2cm}
\caption{
\textbf{Evaluation results on the identification of 20 common visual object categories.} Text in bold indicates the optimal performance for a given category under the $20M$ data setting, while underlined text signifies further performance enhancement achieved when utilizing the entire dataset. All reported scores were generated by GPT-4o on May 6, 2025. To mitigate scoring fluctuations, we conducted three independent scoring iterations and used the mean value as the final score.}
\label{tab:object_result}
\end{table*}

{\bf Text-to-Image Model Training}
We also evaluate the effectiveness of {\ModelName} on the text-to-image generation task. By fine-tuning the Hunyuan-DiT~\citep{li2024hunyuandit}, we aim to demonstrate that our data can improve visual generation capabilities. As evidenced in Table~\ref{tab:t2i_numbers}, models fine-tuned from the publicly released Hunyuan-DiT using our recaption data significantly achieve a lower FID and a higher CLIP score on the MSCOCO dataset and an internal perceptual test set. These findings indicate that our data offer considerable assistance in enhancing the model's understanding of visual concepts and enabling text-to-image models to achieve better performance in terms of authenticity. Realistic cases are presented in Section~\ref{appendix:t2i_case}.

\begin{table}[ht]
\small
\centering	
\resizebox{1.0\columnwidth}{!}{
\begin{tabular}	{ c | c c c | c c c}
\toprule
  & \multicolumn{3}{c}{FID$\downarrow$} & \multicolumn{3}{c}{CLIP Score$\uparrow$} \\
 Dataset & Row Caption & Our Recaption &  Fine-tuning with Our Recaption & Row Caption & Our Recaption  & Fine-tuning with Our Recaption \\
\midrule
perceptive Test Set & 62.38 & 49.13 & {\bf 45.33} &  0.323 & 0.344 & {\bf 0.357} \\
MSCOCO Test set & 28.79 & 21.09 & {\bf 15.46} & 0.307 & 0.305 & {\bf 0.313} \\
 \bottomrule
\end{tabular}
}
\vspace{0.2cm}
\caption
{
{\bf Text-to-Image Evaluation} of the Hunyuan-DiT model on an internal perceptual test set and MSCOCO-30K, comparing performance with and without fine-tuning using our recaption data.
It should be noted that, due to computational resource limitations, we only sampled $2M$ data points from the {\ModelName} dataset for LoRA fine-tuning.
}
\label{tab:t2i_numbers}
\end{table}

\section{Conclusion and Limitation}
\label{sec:conclusion}

This work addresses the challenge of synthesizing low-hallucination, knowledge-rich image captions. We introduce a novel pipeline for generating such high-quality synthetic captions employing two key techniques: continuous DPO (CDPO) and knowledge-enriching SFT. We demonstrate that DPO effectively mitigates hallucinations in captions and analyze its scaling behavior at certain data scales. Furthermore, we present the CDPO approach designed to sustain performance improvements beyond the point where DPO's scaling plateaus.
Experimental results show that {\ModelName} data significantly outperform original alt-text and other synthetic caption datasets for pre-training vision language models. To foster the development of multimodal research within the open source community, we will make {\ModelName} publicly available. See Section~\ref{appendix:limitations} and Section~\ref{appendix:risks} for details on limitations and risks.

\section{Limitations}
\label{appendix:limitations}

Training large vision-language models is inherently computationally intensive and resource-demanding. Specifically, pre-training the 7B-sized VLM model using the complete {\ModelName} dataset required substantial $15,360$ H20 GPU compute hours.
Due to limited computational resources, this work does not explore the full potential of scaling up the use of recaption data. Furthermore, we did not evaluate the effectiveness of using our generated large-scale recaption data to train text-to-image models from scratch. Addressing these limitations by conducting experiments with larger data scales and exploring T2I training from scratch are important avenues for future research.

\section{Potential Risks}
\label{appendix:risks}
The original image-text pairs are primarily derived from open-source datasets. While we have implemented substantial measures to filter out undesirable content, potential risks remain. These risks are particularly salient in the field of image generation, exemplified by issues such as the creation of fake portraits for social media~\citep{hill2020designed}, which is a recognized challenge in this research area.

\bibliographystyle{plainnat}
\bibliography{neurips_2025}

\newpage
\appendix
\section{Loss of Vision-Language Model Training}
\label{appendix:loss}

\begin{figure*}[ht]
\begin{center}
\centerline{\includegraphics[width=1.0\textwidth]{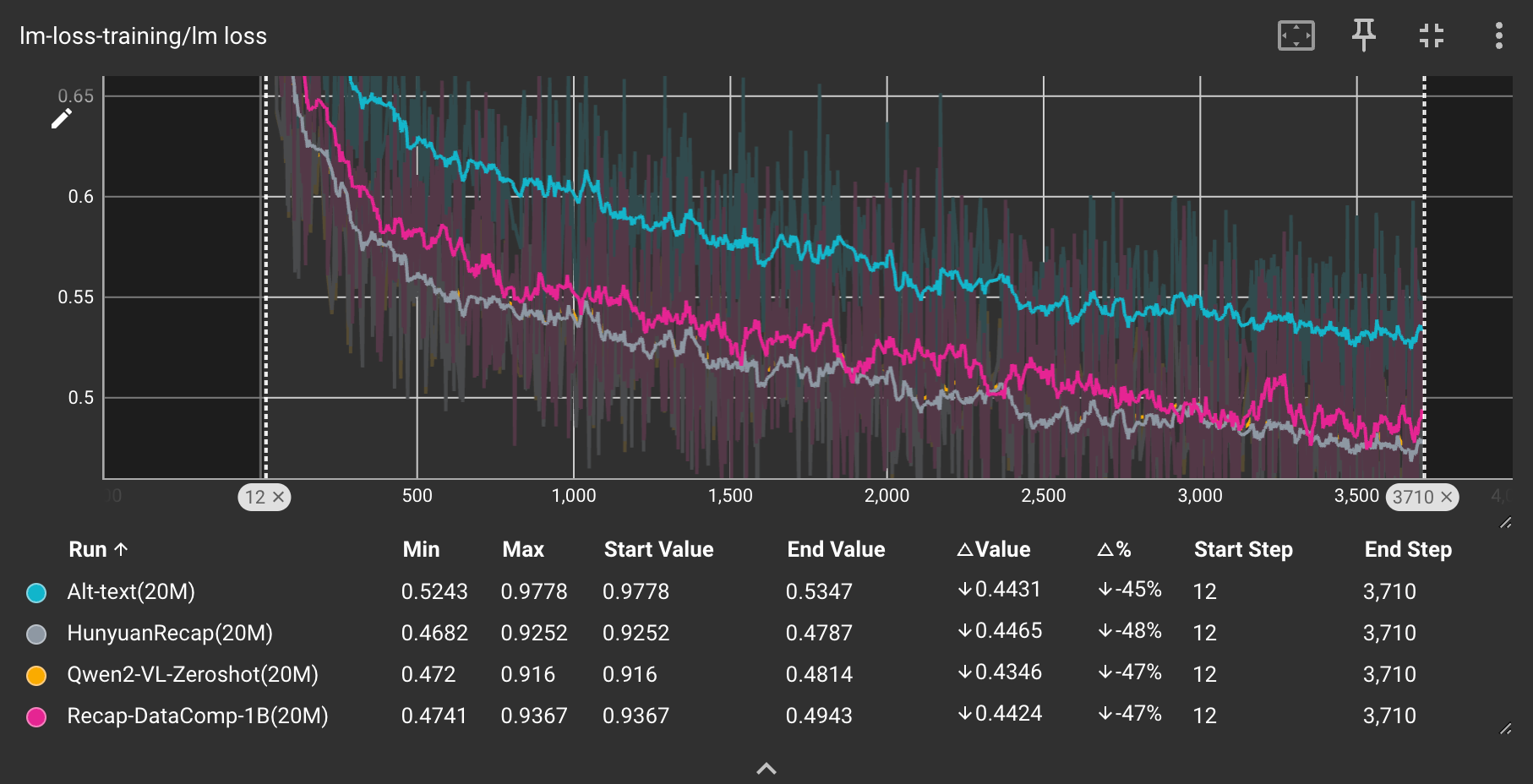}}
\vspace{0.2cm}
\caption{{\bf Loss curve of vision-language model pre-trained with alt-text, Recap-DataComp-1B, Qwen2-VL-Zeroshot or {\ModelName}.} To ensure a fair comparison, all pre-trained models analyzed here utilized the same amount of data ($20M$) for pre-training. Furthermore, the four convergence curves presented are obtained by fine-tuning these models on an identical supervised dataset. }
\label{fig:loss}
\end{center}
\end{figure*}

To address the issue of loss value incomparability arising from diverse pre-training data distributions, we examine the convergence behavior of pre-trained models when fine-tuned on a common supervised dataset~\citep{wang2023cogvlm,li2024llava}, as illustrated in Figure~\ref{fig:loss}. Figure~\ref{fig:loss} shows that models trained with {\ModelName} data exhibit superior convergence. Quantitatively, these models converge to a lower final loss of
$0.4796$ compared to models pre-trained on other datasets.

\section{Examples of Text-to-Image Generations}
\label{appendix:t2i_case}

\begin{figure*}[htbp]
\centering
\begin{spacing}{0.15} 
\setlength\tabcolsep{1pt}
\begin{tabular}{c|cccc}
\toprule
  Text & Image & Hunyuan-DiT & Hunyuan-DiT wRecap & Hunyuan-DiT wFT \\ 
\toprule
  Jellyfish
 & \begin{minipage}[b]{0.22\columnwidth}
    		\centering
    		\raisebox{-.5\height}{\includegraphics[width=\linewidth]{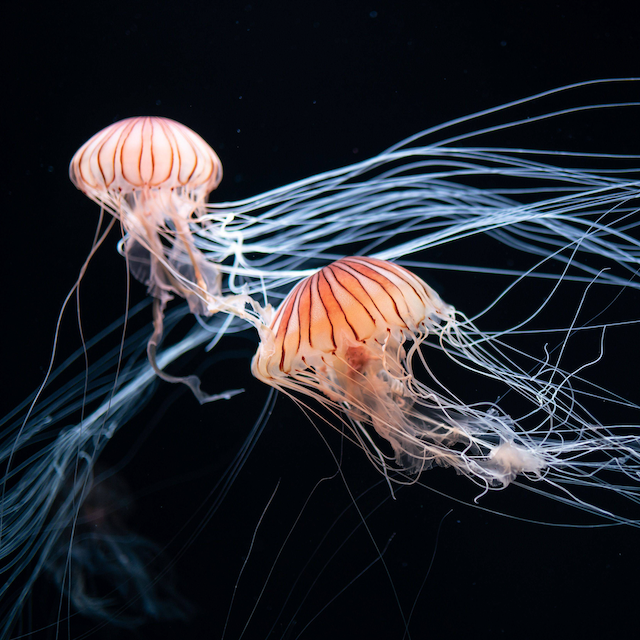}} 
    \end{minipage} & \begin{minipage}[b]{0.22\columnwidth}
    		\centering
    		\raisebox{-.5\height}{\includegraphics[width=\linewidth]{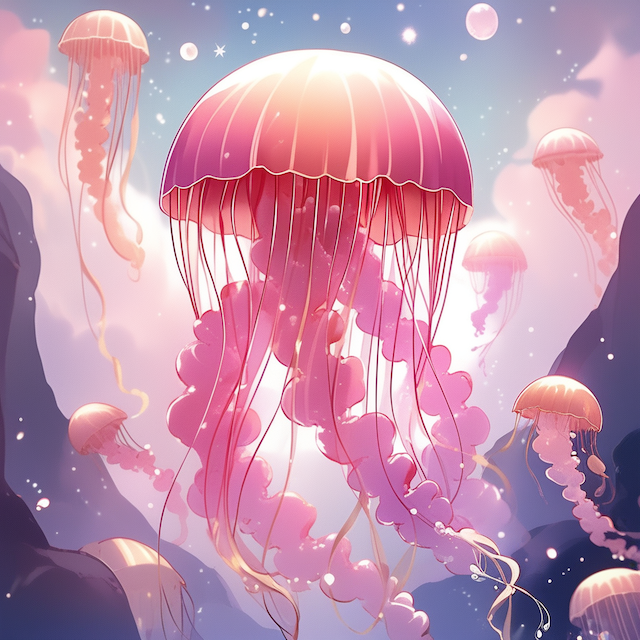}} 
    \end{minipage} & \begin{minipage}[b]{0.22\columnwidth}
    		\centering
    		\raisebox{-.5\height}{\includegraphics[width=\linewidth]{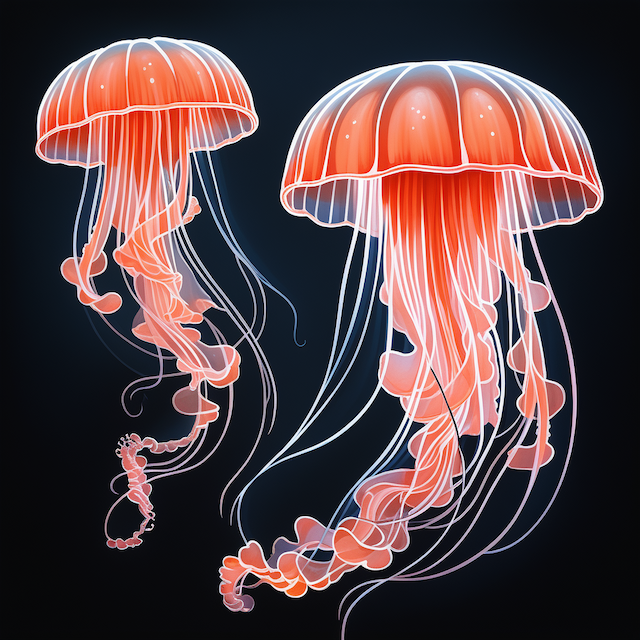}} 
    \end{minipage} & \begin{minipage}[b]{0.22\columnwidth}
    		\centering
    		\raisebox{-.5\height}{\includegraphics[width=\linewidth]{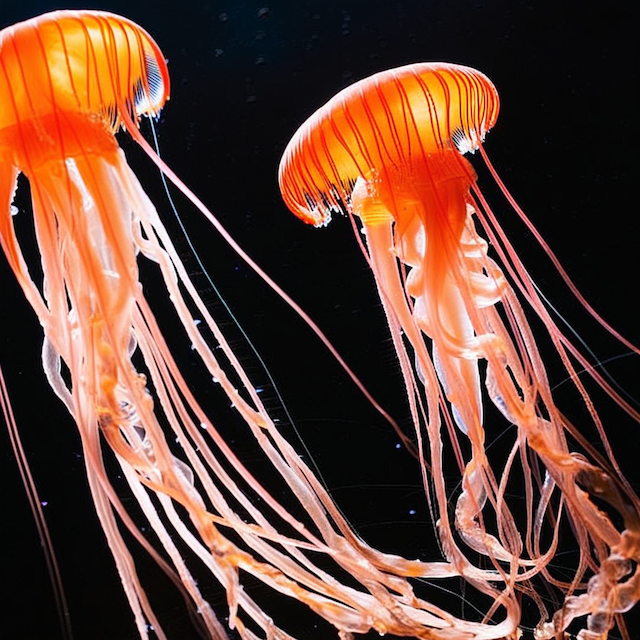}} 
    \end{minipage} \\

  Acerola
 & \begin{minipage}[b]{0.22\columnwidth}
    		\centering
    		\raisebox{-.5\height}{\includegraphics[width=\linewidth]{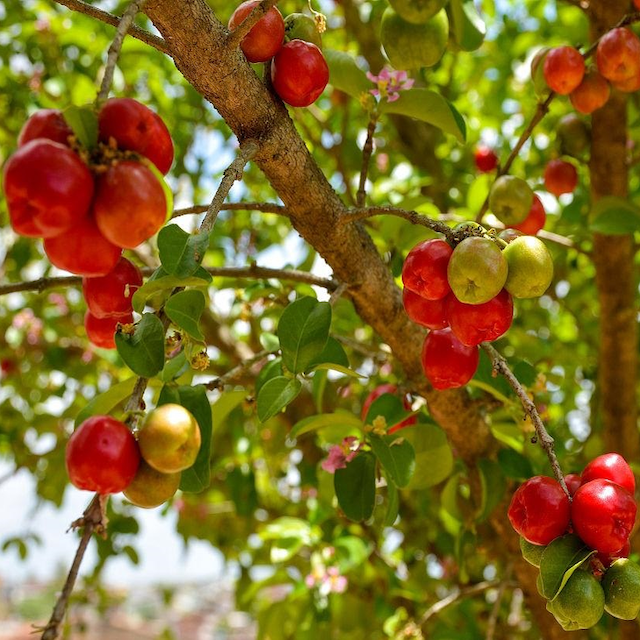}} 
    \end{minipage} & \begin{minipage}[b]{0.22\columnwidth}
    		\centering
    		\raisebox{-.5\height}{\includegraphics[width=\linewidth]{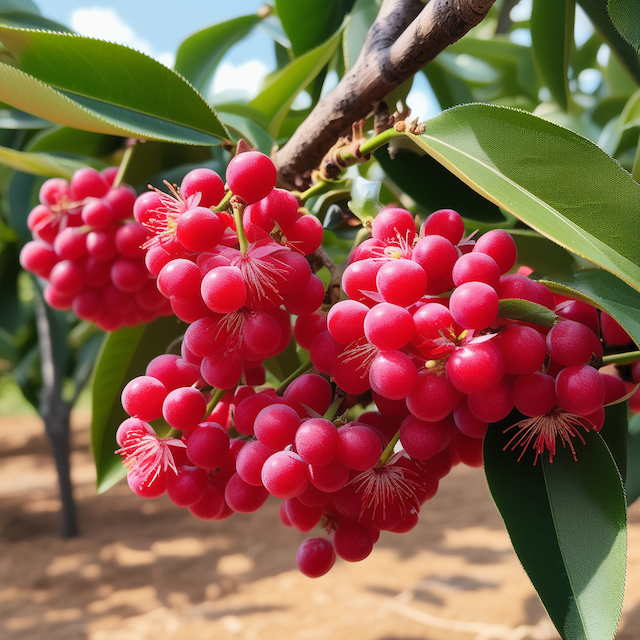}} 
    \end{minipage} & \begin{minipage}[b]{0.22\columnwidth}
    		\centering
    		\raisebox{-.5\height}{\includegraphics[width=\linewidth]{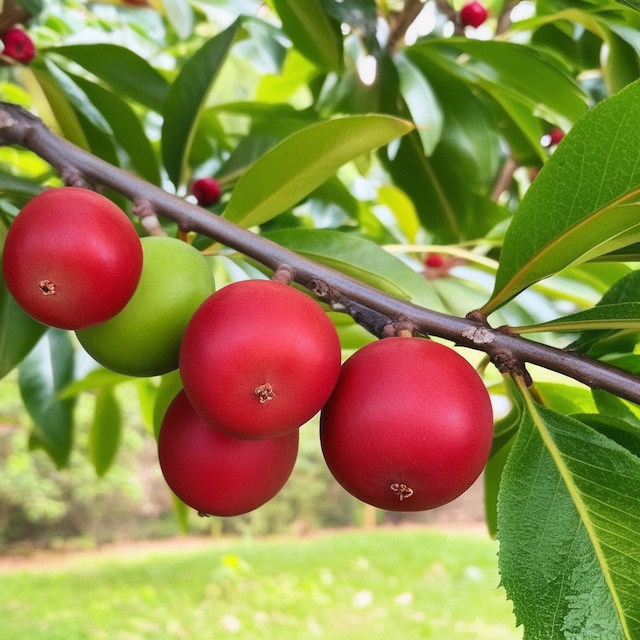}} 
    \end{minipage} & \begin{minipage}[b]{0.22\columnwidth}
    		\centering
    		\raisebox{-.5\height}{\includegraphics[width=\linewidth]{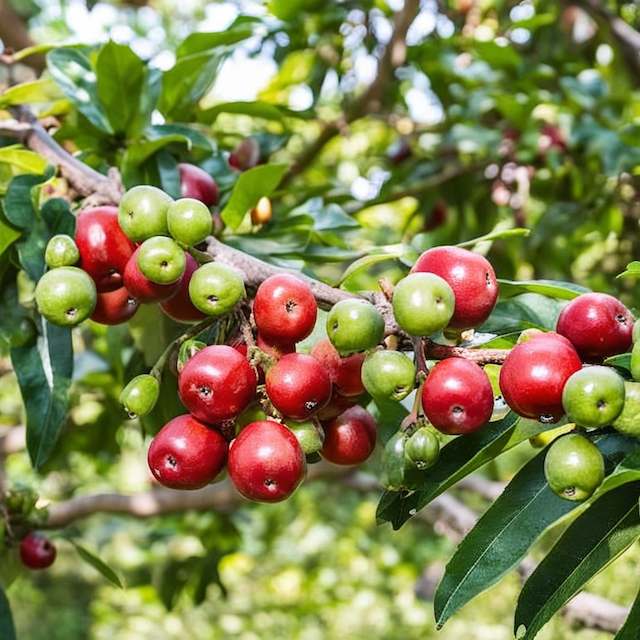}} 
    \end{minipage} \\

  Abacus
 & \begin{minipage}[b]{0.22\columnwidth}
    		\centering
    		\raisebox{-.5\height}{\includegraphics[width=\linewidth]{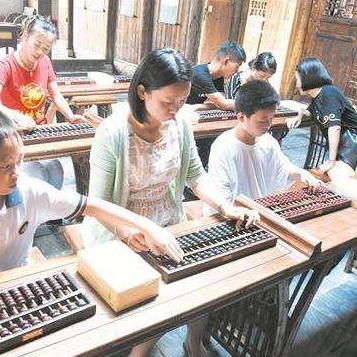}} 
    \end{minipage} & \begin{minipage}[b]{0.22\columnwidth}
    		\centering
    		\raisebox{-.5\height}{\includegraphics[width=\linewidth]{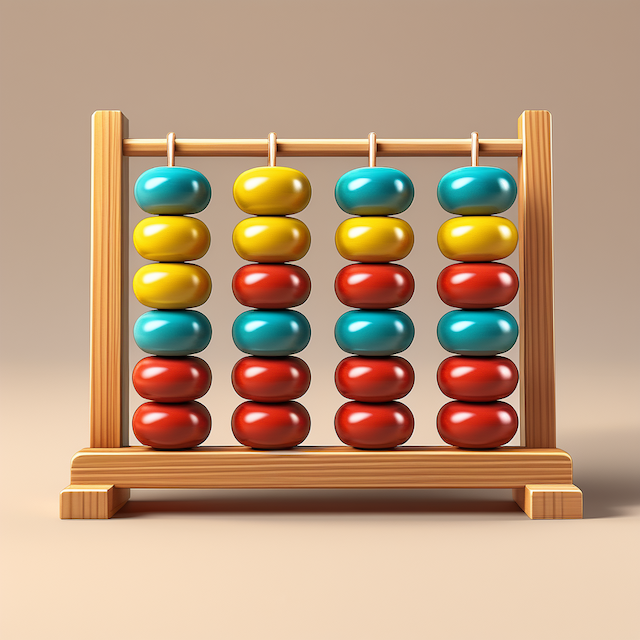}} 
    \end{minipage} & \begin{minipage}[b]{0.22\columnwidth}
    		\centering
    		\raisebox{-.5\height}{\includegraphics[width=\linewidth]{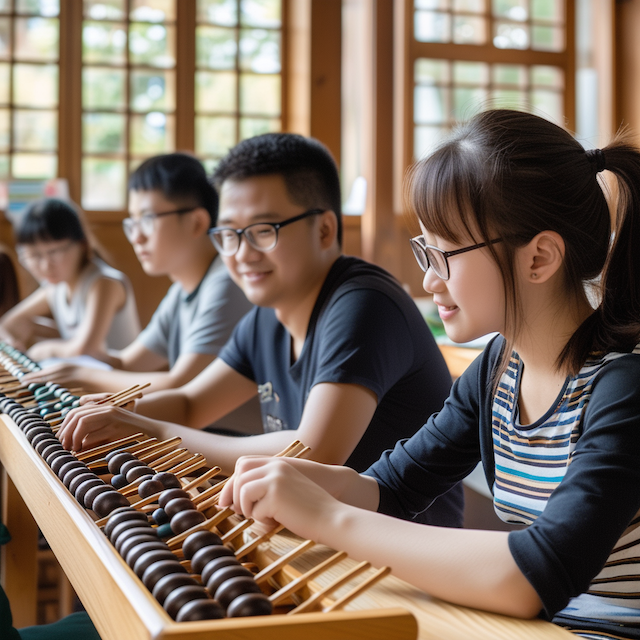}} 
    \end{minipage} & \begin{minipage}[b]{0.22\columnwidth}
    		\centering
    		\raisebox{-.5\height}{\includegraphics[width=\linewidth]{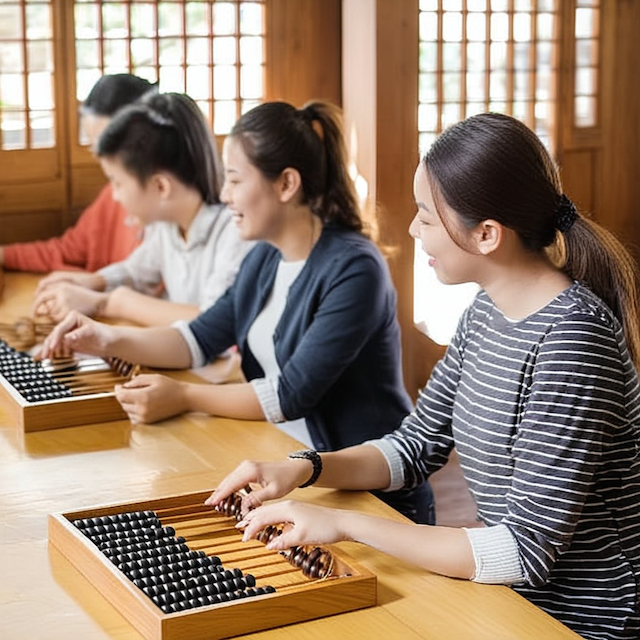}} 
    \end{minipage} \\
  Yueqin
 & \begin{minipage}[b]{0.22\columnwidth}
    		\centering
    		\raisebox{-.5\height}{\includegraphics[width=\linewidth]{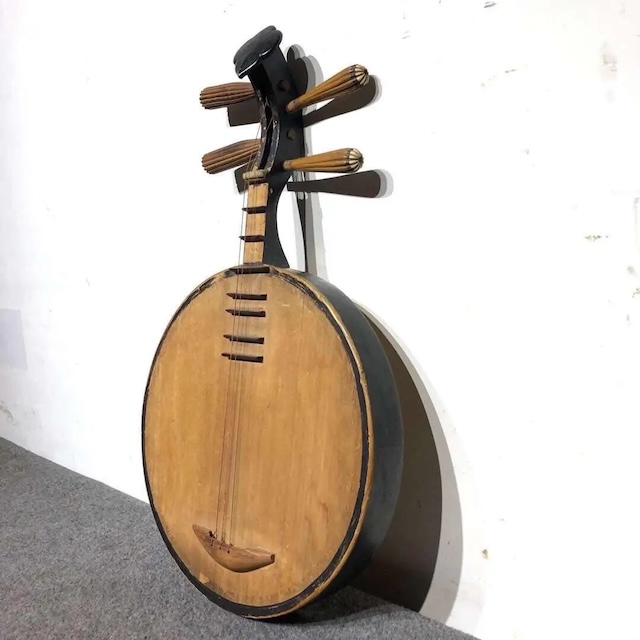}} 
    \end{minipage} & \begin{minipage}[b]{0.22\columnwidth}
    		\centering
    		\raisebox{-.5\height}{\includegraphics[width=\linewidth]{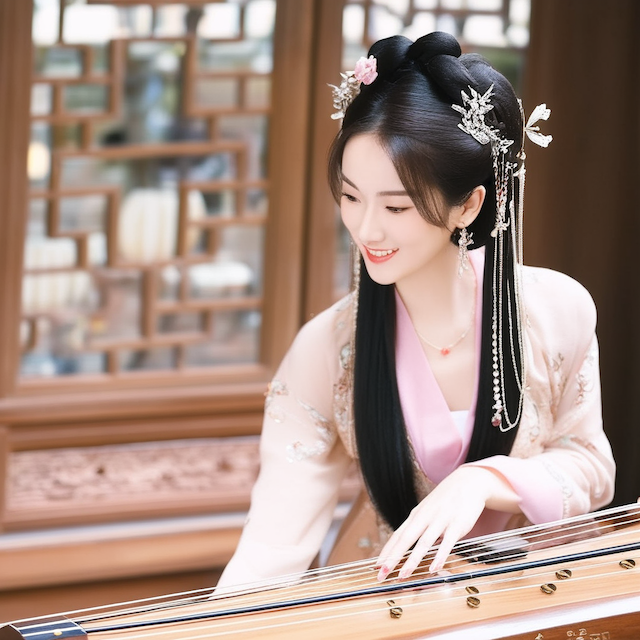}} 
    \end{minipage} & \begin{minipage}[b]{0.22\columnwidth}
    		\centering
    		\raisebox{-.5\height}{\includegraphics[width=\linewidth]{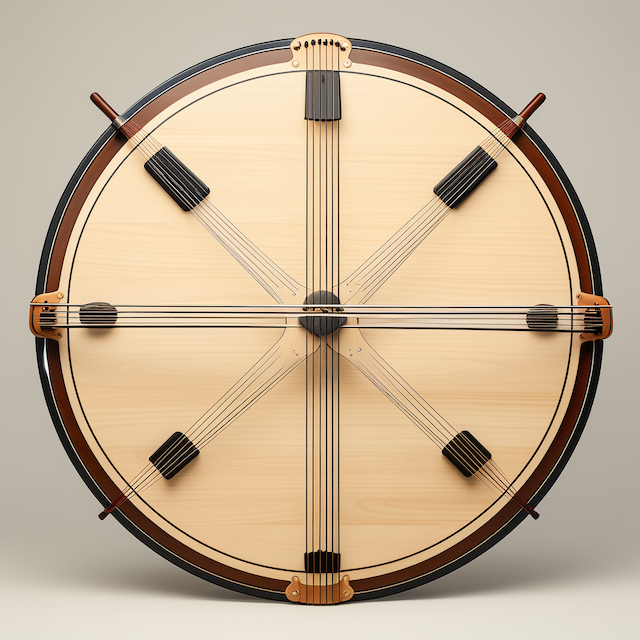}} 
    \end{minipage} & \begin{minipage}[b]{0.22\columnwidth}
    		\centering
    		\raisebox{-.5\height}{\includegraphics[width=\linewidth]{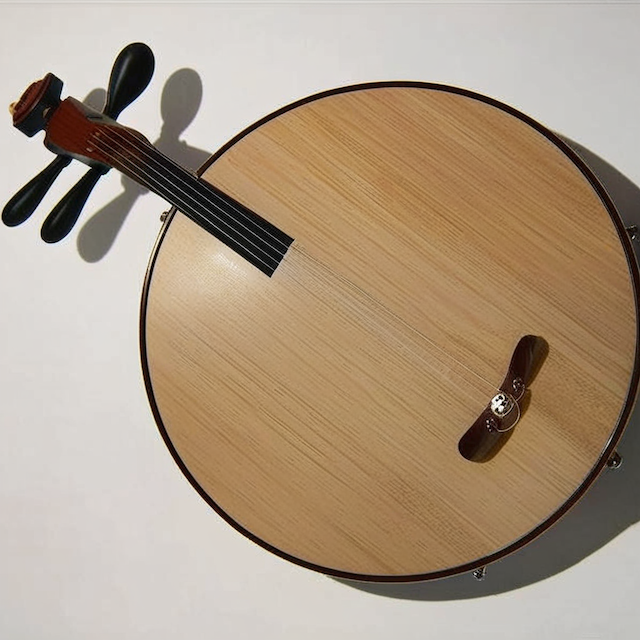}} 
    \end{minipage} \\
  Alpaca
 & \begin{minipage}[b]{0.22\columnwidth}
    		\centering
    		\raisebox{-.5\height}{\includegraphics[width=\linewidth]{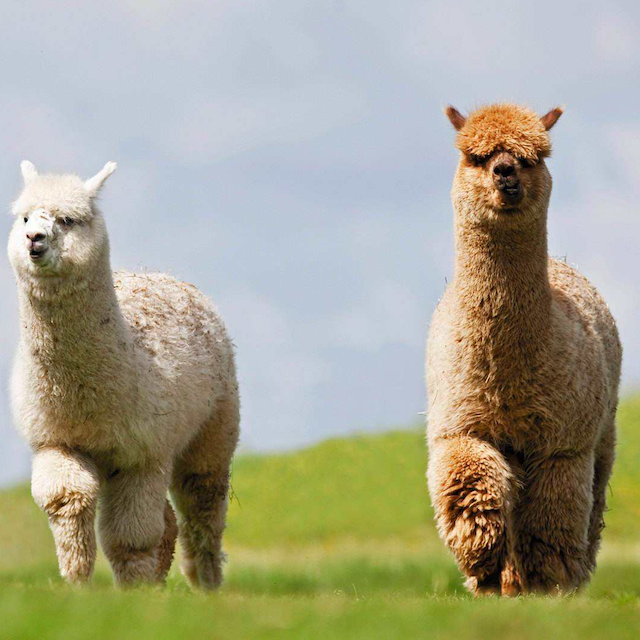}} 
    \end{minipage} & \begin{minipage}[b]{0.22\columnwidth}
    		\centering
    		\raisebox{-.5\height}{\includegraphics[width=\linewidth]{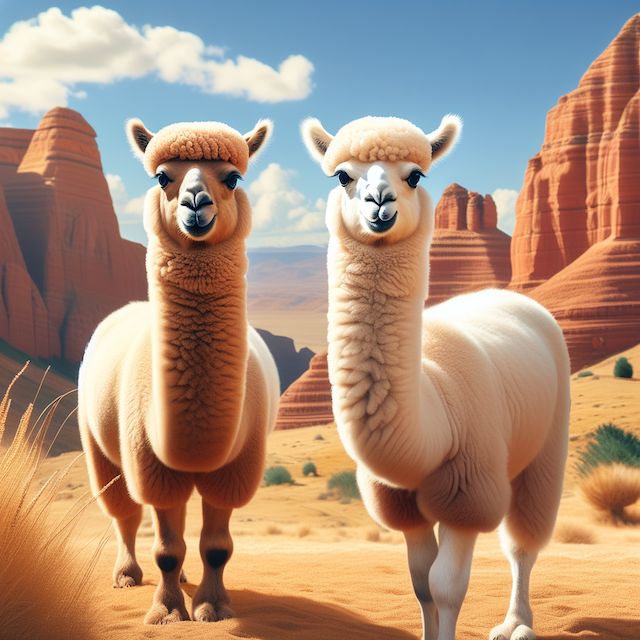}} 
    \end{minipage} & \begin{minipage}[b]{0.22\columnwidth}
    		\centering
    		\raisebox{-.5\height}{\includegraphics[width=\linewidth]{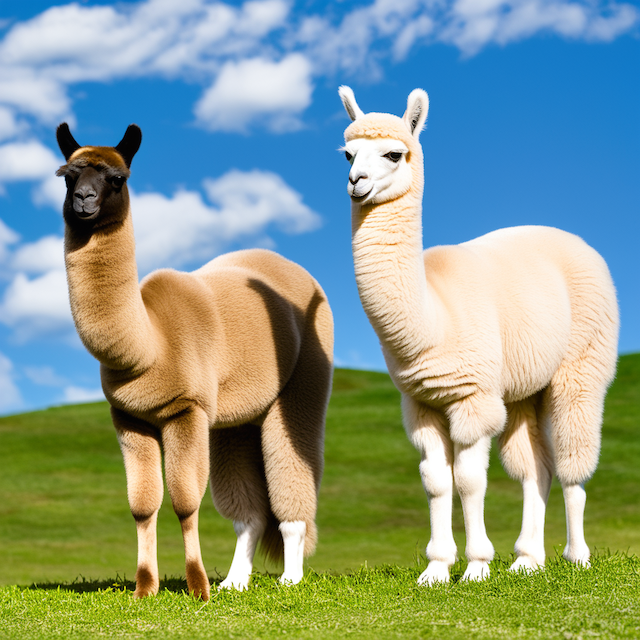}} 
    \end{minipage} & \begin{minipage}[b]{0.22\columnwidth}
    		\centering
    		\raisebox{-.5\height}{\includegraphics[width=\linewidth]{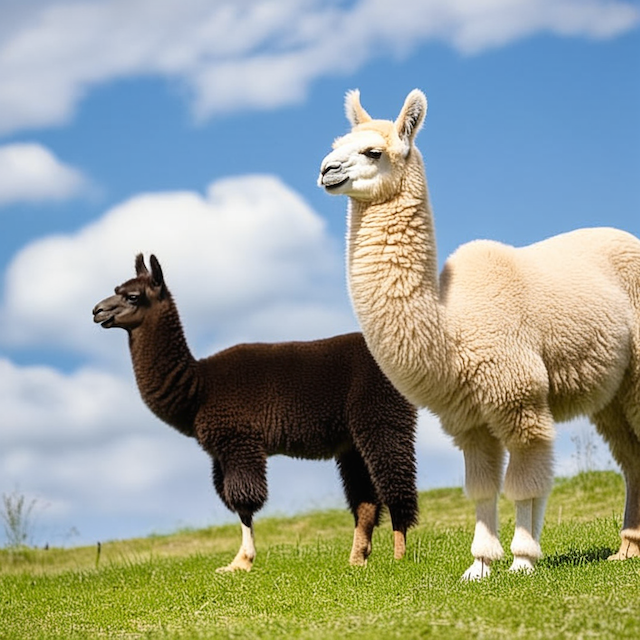}} 
    \end{minipage} \\
  The Grotto
 & \begin{minipage}[b]{0.22\columnwidth}
    		\centering
    		\raisebox{-.5\height}{\includegraphics[width=\linewidth]{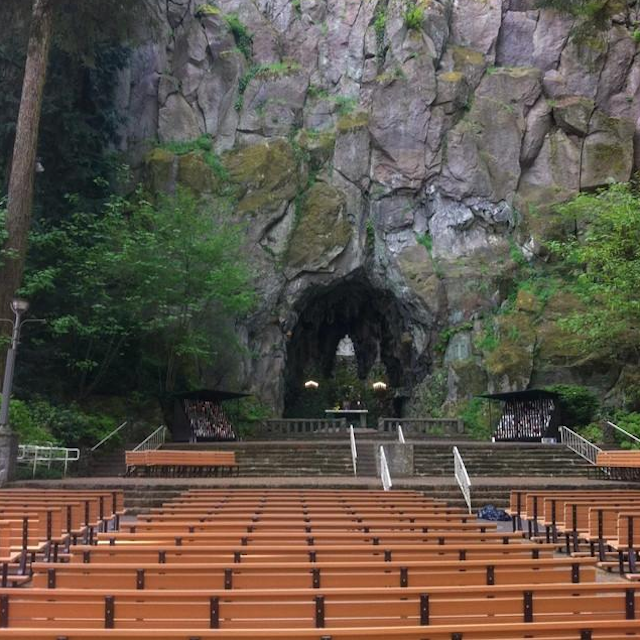}} 
    \end{minipage} & \begin{minipage}[b]{0.22\columnwidth}
    		\centering
    		\raisebox{-.5\height}{\includegraphics[width=\linewidth]{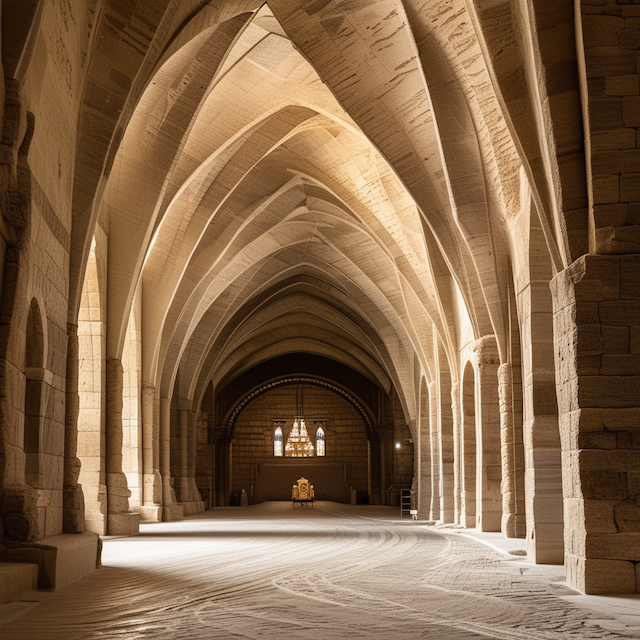}} 
    \end{minipage} & \begin{minipage}[b]{0.22\columnwidth}
    		\centering
    		\raisebox{-.5\height}{\includegraphics[width=\linewidth]{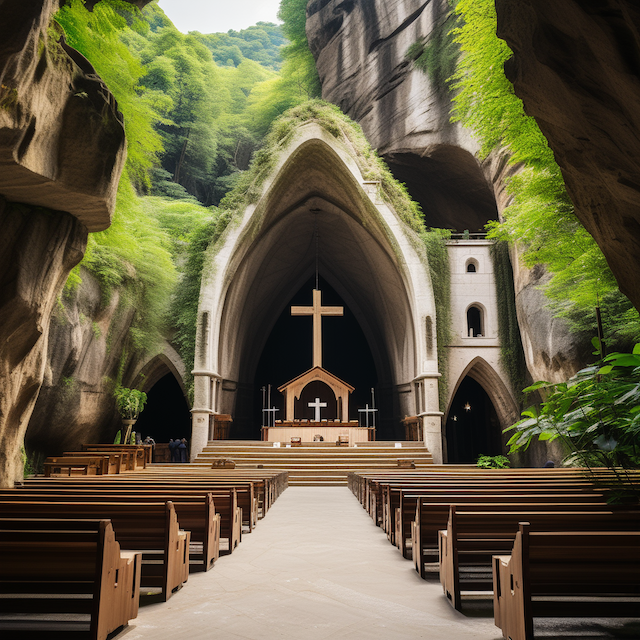}} 
    \end{minipage} & \begin{minipage}[b]{0.22\columnwidth}
    		\centering
    		\raisebox{-.5\height}{\includegraphics[width=\linewidth]{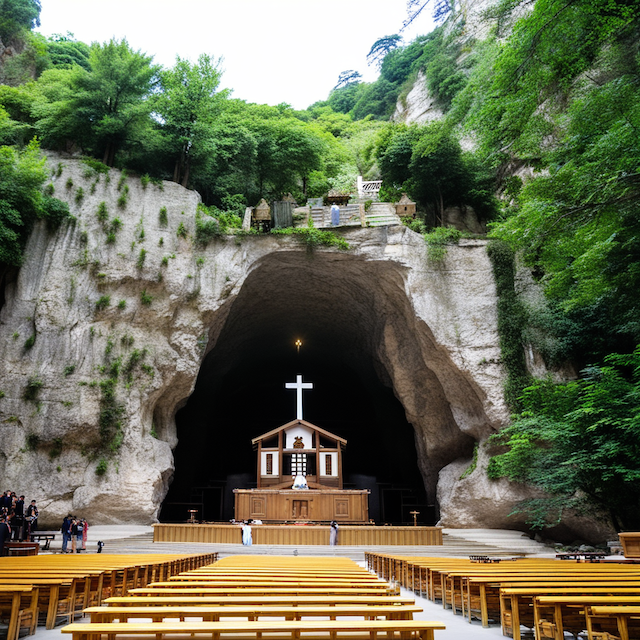}} 
    \end{minipage} \\
  Shanghai Tower
 & \begin{minipage}[b]{0.22\columnwidth}
    		\centering
    		\raisebox{-.5\height}{\includegraphics[width=\linewidth]{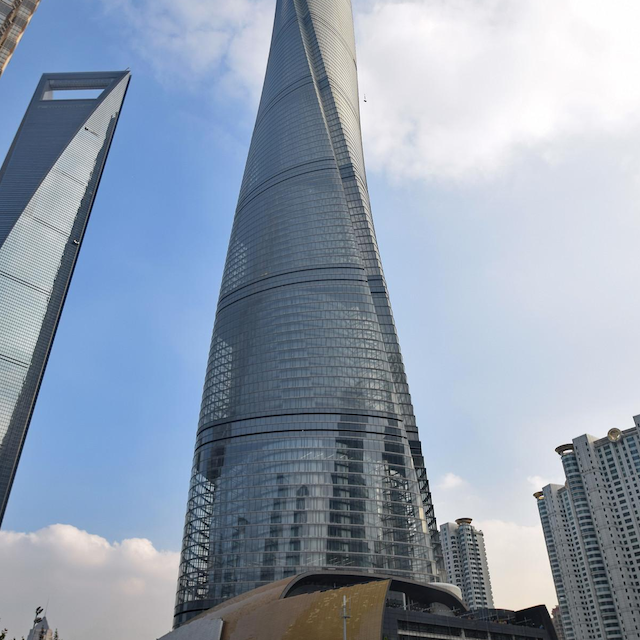}} 
    \end{minipage} & \begin{minipage}[b]{0.22\columnwidth}
    		\centering
    		\raisebox{-.5\height}{\includegraphics[width=\linewidth]{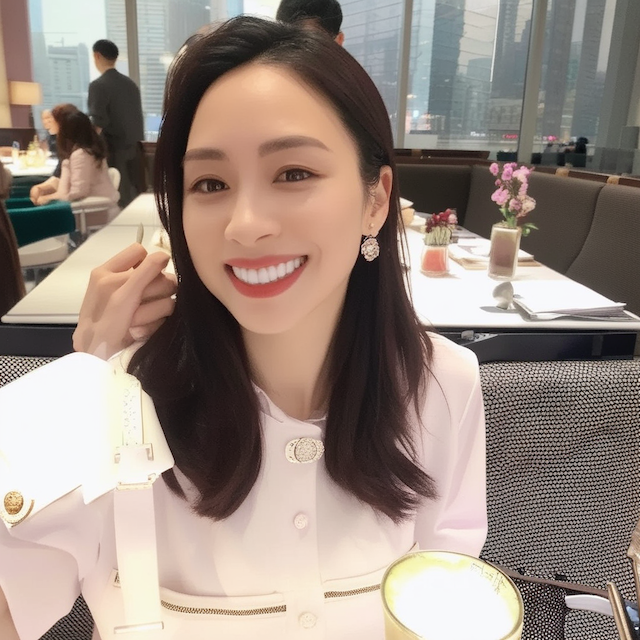}} 
    \end{minipage} & \begin{minipage}[b]{0.22\columnwidth}
    		\centering
    		\raisebox{-.5\height}{\includegraphics[width=\linewidth]{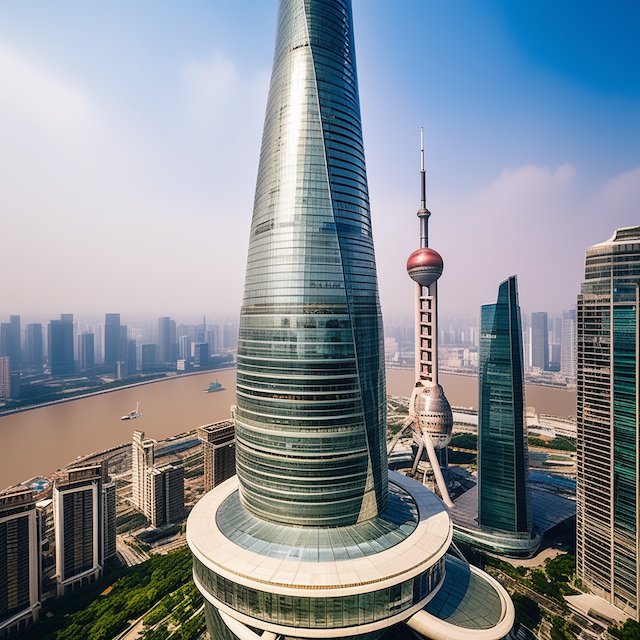}} 
    \end{minipage} & \begin{minipage}[b]{0.22\columnwidth}
    		\centering
    		\raisebox{-.5\height}{\includegraphics[width=\linewidth]{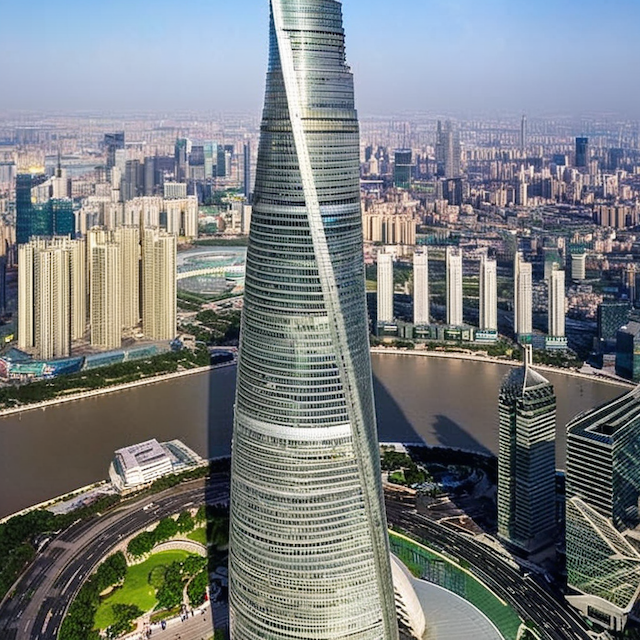}} 
    \end{minipage} \\
    
\bottomrule
\end{tabular}
\end{spacing}
\vspace{0.3cm}
\caption{{\bf Demonstration of text-to-image cases.} Here, the original instruction for image generation is denoted as "Text", the original picture as "Image", the generation effect of the original open-source model as "Hunyuan-DiT", the generation effect of the original open-source model with our Synthetic texts as "Hunyuan-DiT wRecap", and the effect of the model fine-tuned with our recaption data as "Hunyuan-DiT wFT".}
\label{fig:t2i_cases}
\end{figure*}

This section presents examples of text-to-image generation in Figure~\ref{fig:t2i_cases}; the corresponding text prompts are provided in Table~\ref{tab:t2i_recaption}. Observation of these examples reveals that the original Hunyuan-DiT model struggles to accurately render certain concepts (e.g. abacus, Yueqin) and produces less accurate depictions for others (e.g. jellyfish, acerola). However, after fine-tuning with the {\ModelName} data, the model's understanding of these concepts becomes more precise, resulting in generated images with a more realistic style. These qualitative improvements are consistent with the quantitative results shown in Table~\ref{tab:t2i_numbers}, such as the observed reduction in FID. This consistency highlights how the comprehensive concept coverage in our data enhances the model's perceptual fidelity, which is further corroborated by the significant performance improvement of the VLM trained with {\ModelName} on numerous object perception tasks (Table~\ref{tab:object_result}).

\begin{table}[ht]
\centering	
\resizebox{1.0\columnwidth}{!}{
\begin{tabular}	{ c | p{300pt}  }
\toprule
Text & Our Recaption \\
\midrule
Jellyfish & The picture shows two jellyfishes. Their bodies are orange or light red with distinct stripes or lines. Their tentacles are long, slender, transparent or semi-transparent. The background is black, highlighting the colors of the jellyfish and the details of their tentacles. The jellyfish are swimming or floating in the water, and their tentacles are spread out in the water, forming a distinct linear effect.
\\
\midrule
Brazilian Acerola & The image shows the branches of a tree laden with red and green fruits. The red fruits look ripe, while the green ones appear unripe. There are also green leaves on the branches. In the background, more green leaves and a few blurred flowers or small buds can be seen. The overall environment seems to be an outdoor natural setting. \\
\midrule
Abacus & There is a group of people sitting in front of a wooden table in the picture, using abacuses for learning or operation. There are several abacuses on the table. Wooden doors and windows can be seen in the background, indicating that this is an indoor environment, perhaps a classroom or a study place. People are wearing casual or semi-formal clothes, looking relaxed. \\
\midrule
Yueqin & This is a Yueqin, which has a circular resonator and four strings. There are multiple positions for pressing the strings on the body of the instrument, and there are four tuning pegs on the headstock. Overall, it exhibits the characteristics of a traditional musical instrument. \\
\midrule
Alpaca & There are two alpacas standing on the grass in the image, with blue sky and white clouds in the background. The alpaca on the left has lighter fur, while the one on the right has darker fur. The grass is green, and the postures of the alpacas suggest that they are either walking or standing. \\
\midrule
The Grotto & The Grotto is a church located in a natural environment. The facade of the church is a huge rock cave, covered with green vegetation above the cave, and surrounded by dense trees. Inside the church, there is an altar, above which a cross is hung, surrounded by lighting. In front of the church, there is a row of wooden benches for believers to sit and pray. \\
\midrule
Shanghai Tower & Shanghai Tower is a skyscraper with a spiral glass curtain wall design on its exterior. At the top, there is a circular observation deck, surrounded by other high-rise buildings. The overall architectural style is modern and futuristic. \\

 \bottomrule
\end{tabular}
}
\vspace{0.3cm}
\caption
{The original text of the image in Figure~\ref{fig:t2i_cases} and the synthetic caption after being processed by our recaption model.
}
\label{tab:t2i_recaption}
\end{table}


\end{document}